\pdfoutput=1
\PassOptionsToPackage{table}{xcolor}
\documentclass{article} 
\usepackage[final]{colm2026_conference}

\usepackage[utf8]{inputenc}
\usepackage[T1]{fontenc}

\usepackage{microtype}
\usepackage{hyperref}
\usepackage{url}
\usepackage{booktabs}
\usepackage{array}
\usepackage{tabularx}
\usepackage{graphicx}
\usepackage{subcaption}
\usepackage{amsmath}
\usepackage{amssymb}
\usepackage{xspace}
\usepackage{pifont}
\usepackage[most]{tcolorbox}
\usepackage{multirow}

\newcommand{\symbolimg}[2][0.3cm]{%
  \ensuremath{\vcenter{\hbox{\includegraphics[height=#1]{#2}}}}%
}

\newcolumntype{/}{!{\color{white}\vline width 7pt}}

\newcommand{\llama}{\symbolimg[0.35cm]{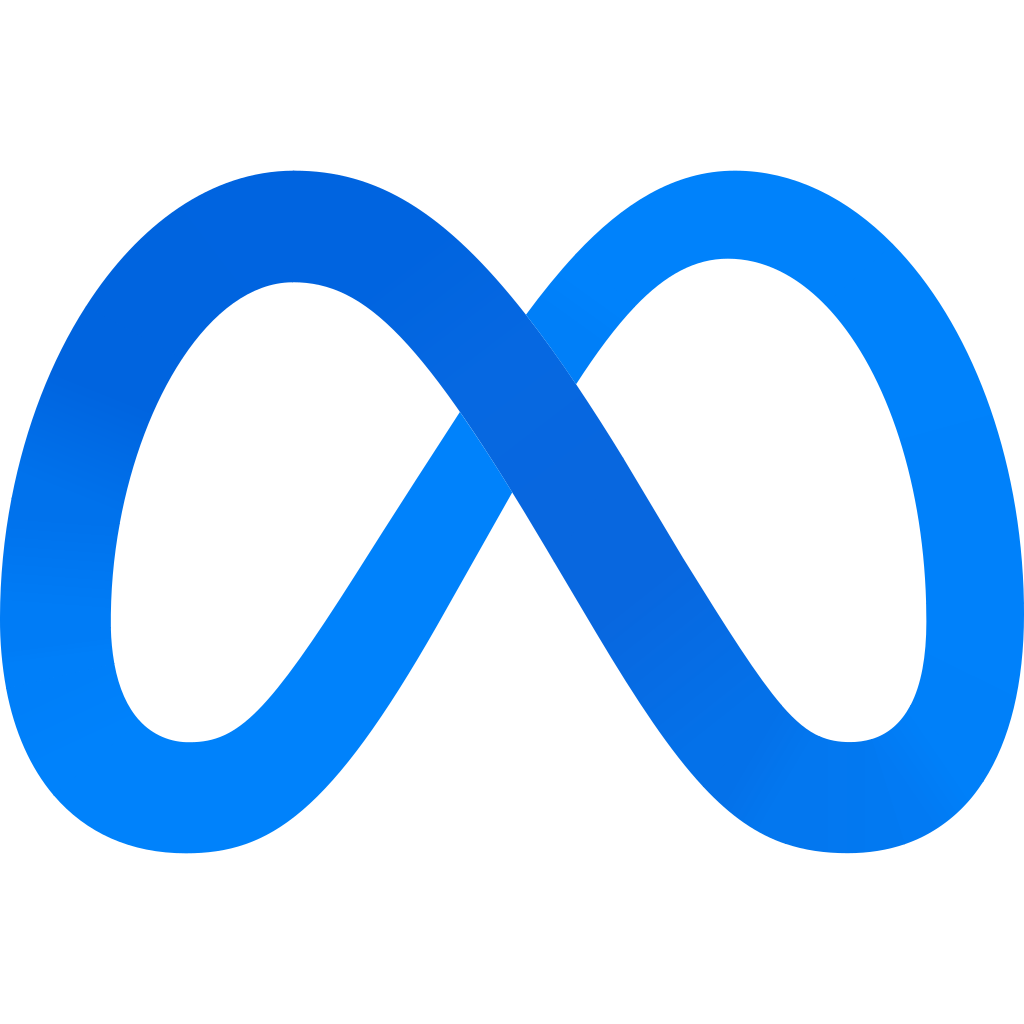}}
\newcommand{\gemini}{\symbolimg[0.35cm]{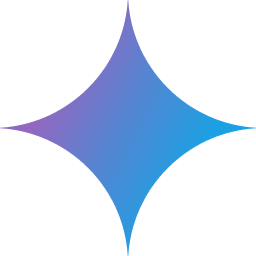}}
\newcommand{\gemma}{\symbolimg[0.35cm]{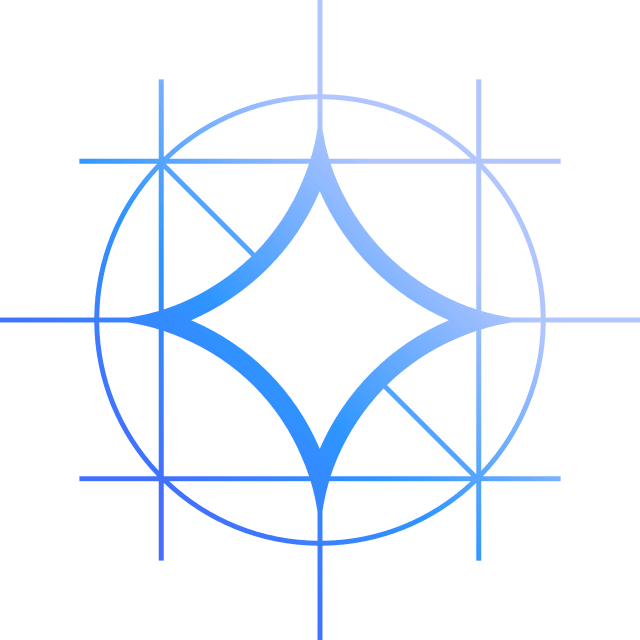}}
\newcommand{\claude}{\symbolimg[0.35cm]{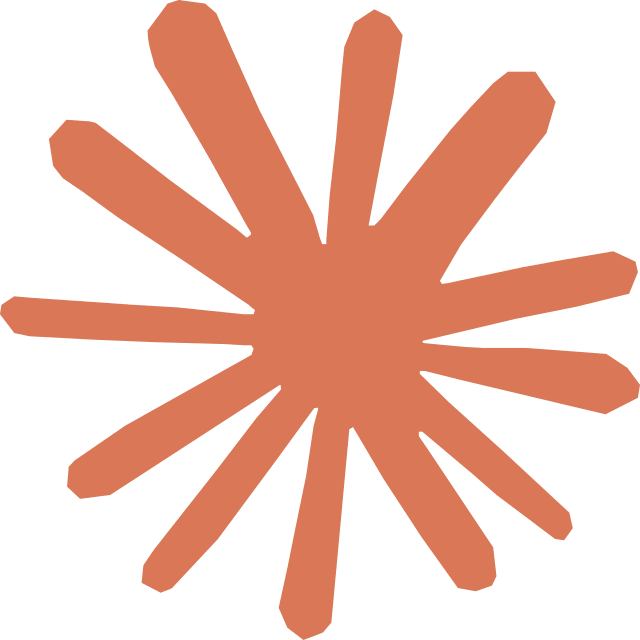}}
\newcommand{\openai}{\symbolimg[0.35cm]{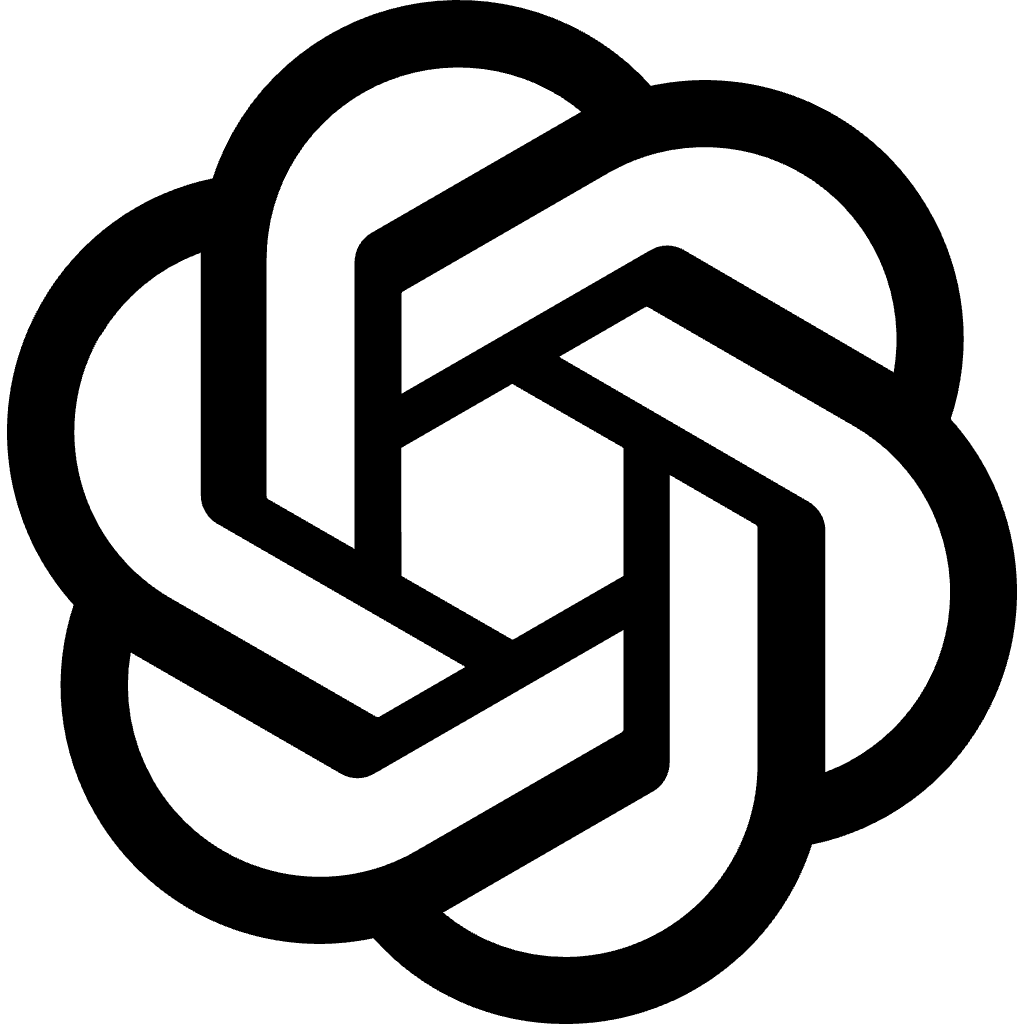}}
\newcommand{\aitwo}{\symbolimg[0.3cm]{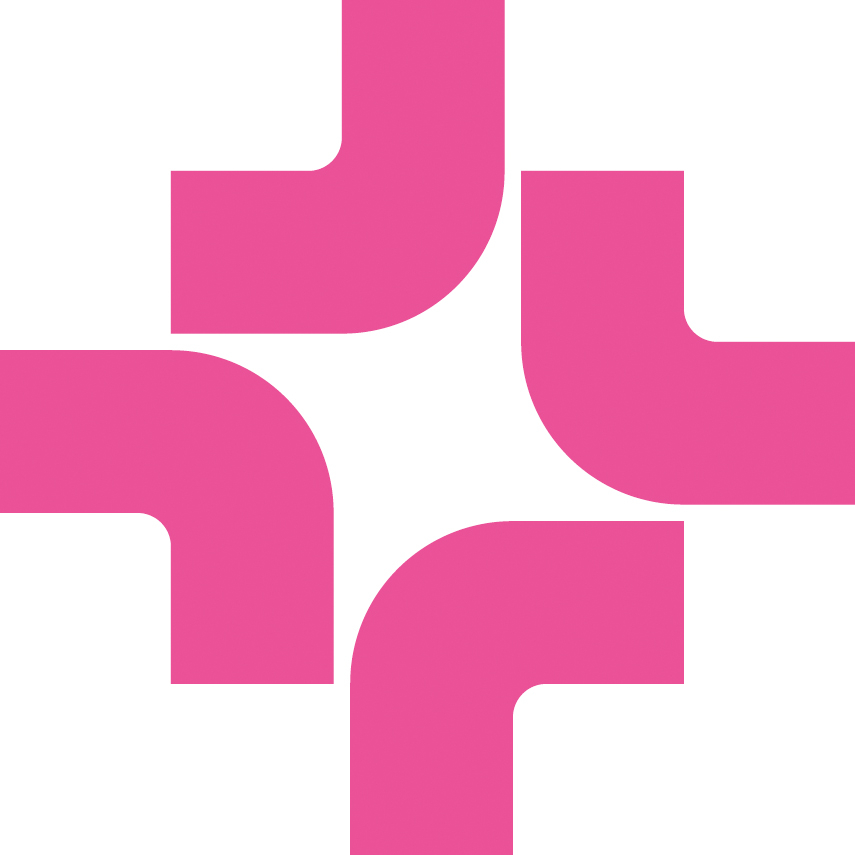}}
\newcommand{\qwen}{\symbolimg[0.35cm]{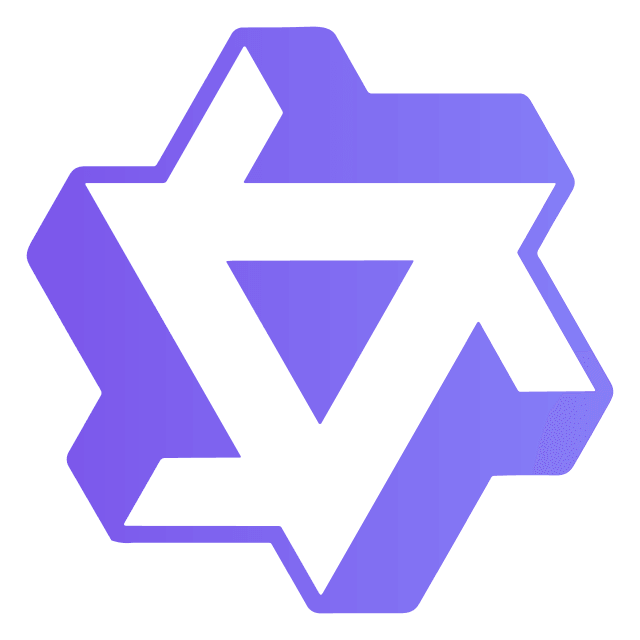}}
\newcommand{\grok}{\symbolimg[0.35cm]{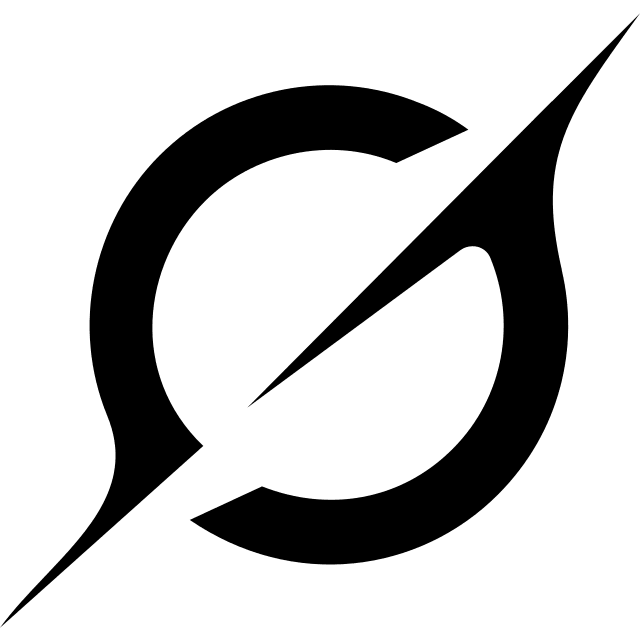}}

\newcommand{\qwenonefive}{\qwen~\textsc{2.5-1.5B-Instruct}\xspace}
\newcommand{\qwenthree}{\qwen~\textsc{2.5-3B-Instruct}\xspace}
\newcommand{\qwenseven}{\qwen~\textsc{2.5-7B-Instruct}\xspace}
\newcommand{\llamaone}{\llama~\textsc{3.2-1B-Instruct}\xspace}
\newcommand{\llamathree}{\llama~\textsc{3.2-3B-Instruct}\xspace}
\newcommand{\llamaeight}{\llama~\textsc{3.1-8B-Instruct}\xspace}
\newcommand{\gemmaone}{\gemma~\textsc{3-1B-it}\xspace}
\newcommand{\gemmafour}{\gemma~\textsc{3-4B-it}\xspace}
\newcommand{\olmothirteen}{\aitwo~\textsc{2-13B-Instruct}\xspace}
\newcommand{\olmonthirtytwo}{\aitwo~\textsc{2-32B-Instruct}\xspace}

\newcommand{\gptfivefourmini}{\openai~\textsc{5.4-mini}\xspace}
\newcommand{\claudehaiku}{\claude~\textsc{Haiku 4.5}\xspace}
\newcommand{\geminiflash}{\gemini~\textsc{2.5-Flash}\xspace}
\newcommand{\grokthreemini}{\grok~\textsc{3-mini}\xspace}

\newcommand{\externalsuite}{\textnormal{\textsc{External}}\xspace}
\newcommand{\historysuite}{\textnormal{\textsc{History}}\xspace}
\newcommand{\iclsuite}{\textnormal{\textsc{ICL}}\xspace}
\newcommand{\ragsuite}{\textnormal{\textsc{RAG}}\xspace}
\newcommand{\toolsuite}{\textnormal{\textsc{Tool}}\xspace}

\definecolor{coolblue}{HTML}{6D9EEB}
\definecolor{warmgold}{HTML}{F1C232}

\newcommand{\cmark}{\ding{51}}

\definecolor{boxframe}{HTML}{1F4E79}
\definecolor{boxbg}{HTML}{F7F7F7}
\definecolor{boxtitlebg}{HTML}{1F4E79}
\definecolor{anchorbg}{HTML}{FFF3CD}
\definecolor{stageonebg}{HTML}{E8F0FE}
\definecolor{stagetwobg}{HTML}{F0F4F0}
\definecolor{labelfg}{HTML}{555555}

\newtcolorbox{examplebox}[1][]{%
  enhanced jigsaw,
  breakable,
  title={Example},
  fontupper=\small,
  #1
}

\newcommand{\anchorhl}[1]{%
  \colorbox{anchorbg}{\strut #1}%
}

\definecolor{darkblue}{rgb}{0, 0, 0.5}
\definecolor{zebrastripe}{RGB}{242,246,252}
\definecolor{champion}{RGB}{242,246,252}
\hypersetup{colorlinks=true, citecolor=darkblue, linkcolor=darkblue, urlcolor=darkblue}

\newcommand{\tablestartstriped}[1]{\rowcolors{#1}{zebrastripe}{white}}
\newcommand{\tablestopstriped}{\rowcolors{1}{white}{white}}

\graphicspath{{figures/}}

\title{AnchorBench: A Multi-Pathway Benchmark for the \\
Anchoring Effect in LLMs}

\author{%
  \normalfont
  \begin{minipage}{\dimexpr\textwidth-2\tabcolsep\relax}
    \centering
    \textbf{Yiderigun Borjigin}\textsuperscript{1} \quad
    \textbf{Alexander Hermann}\textsuperscript{2} \quad
    \textbf{Christian Cyron}\textsuperscript{2,3} \quad
    \textbf{Roland Aydin}\textsuperscript{1,4} \\[0.35em]
    \small
    \textsuperscript{1}Saarland University \quad
    \textsuperscript{2}Hamburg University of Technology \quad
    \textsuperscript{3}Helmholtz-Zentrum Hereon \\
    \textsuperscript{4}German Research Centre for Artificial Intelligence (DFKI)
  \end{minipage}%
}

\begin{document}

\maketitle

\begin{abstract}
  The anchoring effect is a cognitive bias in which an initial reference value shifts a later judgment toward itself. This effect is well established in human judgment and decision-making, and recent work suggests that large language models (LLMs) exhibit similar behavior. However, existing work on anchoring in LLMs typically evaluates only a narrow set of anchor pathways and rarely distinguishes irrelevant from plausible anchors.
We introduce \textsc{AnchorBench}, a benchmark for the anchoring effect in LLMs that evaluates multiple anchor pathways under an explicit anchor relevance axis.
Across fourteen models, including ten open-weight models and four frontier API models, and a large set of controlled prompts, we find that (1) anchoring is strongly pathway-dependent, (2) plausible anchors usually induce larger shifts than irrelevant ones when introduced through stronger pathways, (3) anchor influence generally weakens as the anchor moves farther from the evidence-supported answer, most clearly on External and RAG, and (4) high task accuracy on the anchor-free control condition (Acc$_{10}$: answers within 10 points of gold) does not guarantee robustness: even frontier API models above 95\% control accuracy remain susceptible to plausible anchors.

\end{abstract}

\section{Introduction}
\label{sec:intro}
Large language models (LLMs) are moving beyond text generation into decision-support tasks that require quantitative or evidence-based judgment, including medical question answering \citep{singhalExpertlevelMedicalQuestion2025a} and time-series forecasting \citep{jin2024timellmtimeseriesforecasting,gruver2024largelanguagemodelszeroshot}. In these settings, average accuracy is not enough. A model may give a reasonable answer but still be influenced by a value that appears in the surrounding context, such as an earlier guess in the conversation, a value in a provided example, a retrieved document, or a tool output. This kind of error is especially concerning in high-stakes areas such as medicine. We study this failure mode through the lens of \emph{anchoring}: the tendency for judgments to shift toward a reference value, as introduced by \citet{tversky1974judgment}.

Recent work shows that LLMs exhibit broader human-like cognitive biases across reasoning and decision-making~\citep{jones2022capturing,hagendorff2023humanlike,echterhoff-etal-2024-cognitive,itzhak-etal-2024-instructed}, and more directly that anchoring appears in LLMs~\citep{nguyen2024human,louAnchoringBiasLarge2025,huang2025empirical,takenami-etal-2025-cognitive}.
However, the existing literature still leaves two important gaps.
First, most prior studies test only one or two anchor pathways, usually prompt-embedded anchors.
They rarely compare multiple realistic anchor pathways side by side.
Second, they rarely distinguish clearly between unjustified shifts toward irrelevant anchors and potentially reasonable shifts toward plausible anchors.

\begin{figure*}[t]
  \centering
  \includegraphics[width=\textwidth]{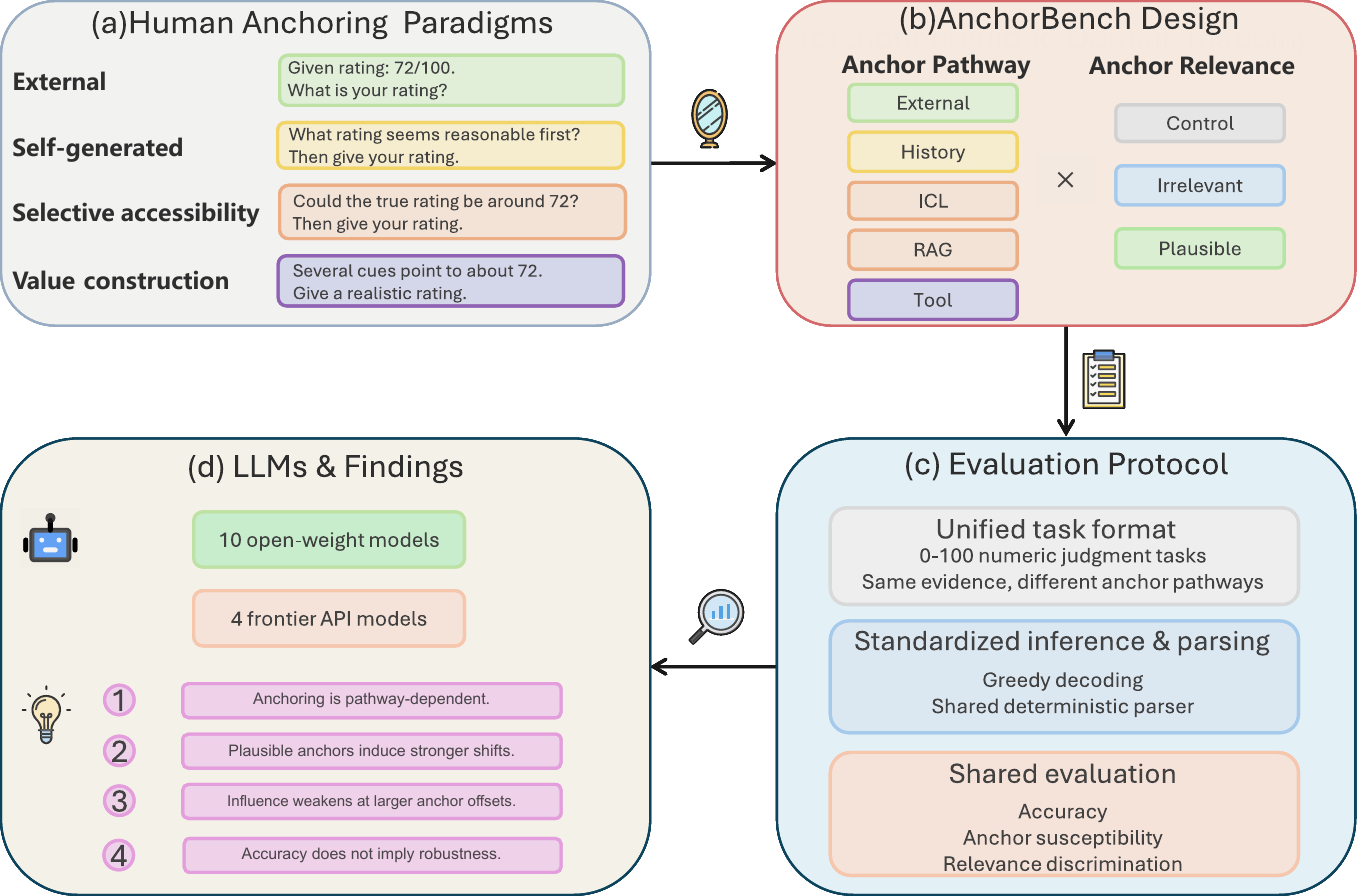}
  \caption{\textsc{AnchorBench} is a unified benchmark for studying anchoring in LLMs, (a) inspired by four human anchoring paradigms, (b) \textsc{AnchorBench} varies anchor pathway and anchor relevance, (c) evaluates them in a shared numeric judgment setup with standardized inference and metrics, (d) compares anchoring behavior across fourteen LLMs.}
  \label{fig:overview}
\end{figure*}

We address these gaps with \textsc{AnchorBench},\footnote{Code and data: \url{https://github.com/Ydrg9989/AnchorBench}.} a controlled diagnostic for the anchoring effect in LLMs.
As shown in Figure~\ref{fig:overview}, the benchmark covers five anchor pathways that correspond to standard ways context reaches a deployed model: External (the prompt), History (conversation), In-Context Learning (ICL, demonstrations), Retrieval-Augmented Generation (RAG, retrieved documents), and Tool (tool outputs).
These pathways also have loose parallels in the human anchoring literature~\citep{tversky1974judgment,epley2001putting,mussweiler1999hypothesis,chapman1999anchoring}, which we use as informal motivation rather than strict one-to-one equivalences.
It also introduces an explicit anchor relevance axis with control, irrelevant, and plausible conditions.
Because every item contains structured numeric evidence and a deterministic gold answer, the benchmark measures not only whether outputs shift, but also whether the shift is justified.

Across the experiments, we find that (1) anchoring is strongly pathway-dependent, (2) plausible anchors usually produce larger shifts than irrelevant anchors when the pathway is strong, (3) anchor influence weakens as the anchor moves farther from the underlying evidence, and (4) accuracy and robustness come apart: even frontier API models with very high control-condition accuracy (Acc$_{10}$: answers within 10 points of gold) remain measurably susceptible to anchoring, though at smaller magnitudes than open-weight models.

Our contributions are as follows:
\begin{enumerate}
  \item \textbf{A multi-pathway diagnostic benchmark for LLMs.} We design five realistic anchor delivery pathways that mirror how context reaches a deployed model, loosely informed by classic human anchoring theory, and evaluate them under a shared framework.
  \item \textbf{A relevance-aware evaluation design.} We separate control, irrelevant, and plausible anchors. This lets us distinguish unambiguous bias (any shift toward an irrelevant anchor, which carries no task-relevant information) from sensitivity to plausible anchors, where a bounded shift can be consistent with rational evidence integration but a sufficiently large shift cannot.
  \item \textbf{A broad empirical comparison across model families and access regimes.} We evaluate fourteen models, including ten open-weight models and four frontier API models, across five suites and 9,000 condition-controlled prompts per model.
\end{enumerate}

\section{Related work}
\label{sec:related}
\paragraph{Human anchoring theory.}
Anchoring is a classic finding in judgment and decision-making: estimates are often drawn toward an initial value, even when that value is arbitrary or only weakly informative~\citep{tversky1974judgment}.
No single mechanism explains it. For externally provided anchors, selective-accessibility accounts argue that people test anchor-consistent hypotheses and retrieve anchor-consistent knowledge~\citep{strack1997explaining,mussweiler1999hypothesis}; for self-generated anchors, anchoring-and-adjustment accounts emphasize insufficient adjustment from an internal starting point~\citep{epley2001putting}; and value-construction accounts tie anchoring to how values are activated and assembled during judgment~\citep{chapman1999anchoring}. Anchor effectiveness also depends on plausibility and extremity~\citep{wegener2001implications,furnham2011literature}.
The human effect is sizable and relevance-dependent: \citet{pIndividualDifferencesAnchoring2019} reports Cohen's $d$ from 0.14 to 1.00 across 24 items, and \citet{LI2021101629} finds stronger anchoring for related than random anchors---a reference scale we revisit in Appendix~\ref{app:cohens-d}.
These distinctions motivate our design: some pathways correspond to classic external or self-generated anchoring, while others operationalize selective-accessibility- or value-construction-style influence through demonstrations, retrieval, and tools.

\paragraph{Anchoring effect in LLMs.}
Anchoring also appears in LLMs. Frontier models shift their estimates toward previously mentioned values, and prompt-based mitigations such as chain-of-thought, reflection, or explicit instructions to ignore the anchor give only limited and inconsistent relief~\citep{nguyen2024human,louAnchoringBiasLarge2025}; the effect persists in multi-turn price negotiation, where reasoning models are less susceptible~\citep{takenami-etal-2025-cognitive}.
\textsc{SynAnchors}~\citep{huang2025empirical} is the closest prior benchmark and is complementary: it varies the \emph{magnitude} of in-prompt anchors and uses causal tracing, whereas \textsc{AnchorBench} fixes the numeric value and varies its \emph{framing} (irrelevant vs.\ plausible) and \emph{delivery pathway}. Where they overlap, our findings agree.
Most of this work studies one pathway at a time, which leaves open how External, History, ICL, RAG, and Tool anchoring compare within a unified setup.

\paragraph{Cognitive bias in LLMs.}
Anchoring is one of several human cognitive biases documented in LLMs, which have been used as a lens on systematic model failure~\citep{jones2022capturing,hagendorff2023humanlike}, studied as an emergent effect of instruction tuning~\citep{itzhak-etal-2024-instructed}, and observed in decision-making and in survey response, where models are unreliable stand-ins for human respondents~\citep{echterhoff-etal-2024-cognitive,tjuatja-etal-2024-llms}; broader evaluations and surveys catalogue many such biases and their mitigations~\citep{malberg-etal-2025-comprehensive,sumitaCognitiveBiasesLarge2024}.
The pattern is especially well documented in LLM-as-a-judge pipelines, where position and related judging biases distort model-based evaluation~\citep{koo-etal-2024-benchmarking,zheng2023judgingllmasajudgemtbenchchatbot,wang-etal-2024-large-language-models-fair,ye2024justiceprejudicequantifyingbiases,wang2025assessing} and evaluators show anchoring in multi-attribute scoring~\citep{stureborg2024largelanguagemodelsinconsistent}---so the effect reaches evaluation pipelines, not only end-user tasks.

\paragraph{Context influence beyond anchoring.}
Numerical anchoring is one instance of a broader sensitivity to non-informative context, alongside sycophancy and persuasion, where models shift toward a user's view or a confidently framed claim~\citep{sharma2025understandingsycophancylanguagemodels,ranaldi2025largelanguagemodelscontradict} and toward authoritative-sounding sources~\citep{anagnostidis2024susceptiblellmsinfluenceprompts,zhou2023navigatinggreyareaexpressions,nguyen-etal-2025-persuasive}.
We see anchoring as narrower: it isolates a single numeric value rather than opinions or arguments, which lets us measure the shift against a deterministic gold answer.

\section{Benchmark setup and evaluation design}
\label{sec:benchmark}
\textsc{AnchorBench} is a controlled diagnostic rather than a sample of organic user queries: every item is synthetic and built so that the gold answer is fixed by the evidence alone, which lets us measure anchoring as a deviation from a known target.
It combines two controlled design factors: the anchor pathway, which determines how an anchor reaches the model, and anchor relevance, which determines how the anchor is framed relative to the task.

\subsection{Item construction}
\label{sec:item-construction}

Each item is a numeric judgment task on a shared 0--100 scale across
six domains (pricing, operations, logistics, resource consumption,
market adoption, compliance).
As illustrated in Figure~\ref{fig:main-item-format}a, the model
receives a set of numeric \emph{evidence ratings} and must return a
single integer estimate.
The gold answer is the rounded arithmetic mean of the visible ratings:
\begin{equation}
  y^{*} \;=\; \mathrm{round}\!\Bigl(\tfrac{1}{k}
  \textstyle\sum_{j=1}^{k} e_j\Bigr),
\end{equation}
where $e_1,\ldots,e_k$ are the ratings shown to the model.
This makes the task a deterministic aggregation problem:
a model that correctly averages the given numbers matches gold exactly.

Item difficulty is governed by two parameters.
A latent center $\theta \sim \mathrm{Unif}[30,70]$ sets the evidence
range; ratings are sampled as $e_j \sim \mathcal{N}(\theta,\sigma^2)$,
clipped to $[0,100]$.
\textbf{Easy} items present all five ratings with low noise
($\sigma{=}8$).
\textbf{Hard} items hide two of the five ratings, add a conflicting
value, and raise noise ($\sigma{=}15$), leaving fewer and less
consistent signals for aggregation.
The six core domains are business-oriented, but a pilot adding
medical, legal, and consumer domains shows the same
UAI$_{\text{pls}}>$UAI$_{\text{irr}}>0$ pattern
(Appendix~\ref{app:extension}).

\paragraph{Provenance of the items.}
Items come from author-designed templates: scenarios, evidence
labels, questions, and anchor framings are drawn from pools
generated with LLM assistance and verified by the authors, while
the numeric evidence, anchor values, and gold answers are produced
by seeded sampling, never by a model.
Holding the numeric value and its position fixed while varying only
the framing sentence is the matched control the relevance axis
needs, and organic queries cannot supply it---at a cost in external
validity we state rather than claim away.

\definecolor{lightgraybox}{RGB}{245,245,245}
\definecolor{grayframe}{RGB}{170,170,170}

\definecolor{lightbluebox}{RGB}{237,244,252}
\definecolor{blueframe}{RGB}{126,166,214}

\definecolor{notetext}{RGB}{50,50,50}
\definecolor{slottext}{RGB}{95,95,95}

\begin{figure}[t]
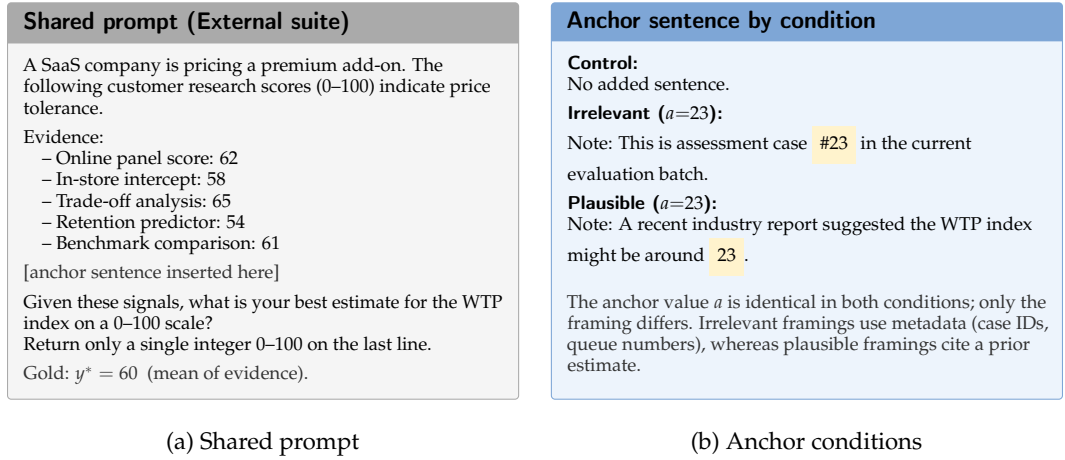

  \centering
  \begin{subfigure}[t]{0.485\columnwidth}
    \begin{tcolorbox}[
        colback=lightgraybox,
        colframe=grayframe,
        boxrule=0.4pt,
        arc=1.2pt,
        left=3pt, right=3pt, top=2.5pt, bottom=2.5pt,
        title={\footnotesize\textbf{Shared prompt (External suite)}},
        fonttitle=\footnotesize\sffamily,
        coltitle=black,
        enhanced,
        equal height group=main-item-boxes,
        valign=top,
        before upper=\raggedright,
      ]
      \scriptsize
      A SaaS company is pricing a premium add-on.
      The following customer research scores (0--100) indicate
      price tolerance.

      \smallskip
      Evidence:\par
      {\scriptsize
        \hspace{1em}-- Online panel score: 62\par
        \hspace{1em}-- In-store intercept: 58\par
        \hspace{1em}-- Trade-off analysis: 65\par
        \hspace{1em}-- Retention predictor: 54\par
      \hspace{1em}-- Benchmark comparison: 61}

      \smallskip
      {\color{notetext}\scriptsize [anchor sentence inserted here]}

      \smallskip
      Given these signals, what is your best estimate
      for the WTP index on a 0--100 scale?\par
      Return only a single integer 0--100 on the last line.

      \smallskip
      {\color{notetext}\scriptsize Gold: $y^{*} = 60$\enspace
      (mean of evidence).}
    \end{tcolorbox}
    \caption{Shared prompt}
  \end{subfigure}
  \hfill
  \begin{subfigure}[t]{0.485\columnwidth}
    \begin{tcolorbox}[
        colback=lightbluebox,
        colframe=blueframe,
        boxrule=0.4pt,
        arc=1.2pt,
        left=3pt, right=3pt, top=2.5pt, bottom=2.5pt,
        title={\footnotesize\textbf{Anchor sentence by condition}},
        fonttitle=\footnotesize\sffamily,
        coltitle=black,
        enhanced,
        equal height group=main-item-boxes,
        valign=top,
        before upper=\raggedright,
      ]
      \scriptsize
      {\sffamily\bfseries Control:}\par
      No added sentence.

      \smallskip
      {\sffamily\bfseries Irrelevant ($a{=}23$):}\par
      Note: This is assessment case \anchorhl{\#23}
      in the current evaluation batch.

      \smallskip
      {\sffamily\bfseries Plausible ($a{=}23$):}\par
      Note: A recent industry report suggested the WTP index
      might be around \anchorhl{23}.

      \medskip
      {\color{notetext}\scriptsize
        The anchor value $a$ is identical in both conditions; only the framing differs.
        Irrelevant framings use metadata (case IDs, queue numbers), whereas
        plausible framings cite a prior estimate.}
    \end{tcolorbox}
    \caption{Anchor conditions}
  \end{subfigure}
  \caption{Example \textsc{AnchorBench} item (External suite, pricing domain).
    All conditions share the same scenario, evidence, question, and answer instruction.
    Only the anchor sentence differs. Additional examples are in
  Appendix~\ref{app:example-boxes}.}
  \label{fig:main-item-format}
\end{figure}

\subsection{Anchor conditions and relevance axis}
\label{sec:anchor-conditions}

Each item is presented under five matched conditions, illustrated
in Figure~\ref{fig:main-item-format}b.
The \emph{control} condition presents the task without any anchor.
The four \emph{anchored} conditions cross two relevance
levels---\emph{irrelevant} and \emph{plausible}---with a low and a
high anchor direction each, yielding:
\texttt{control},
\texttt{irrelevant\_low/high}, and \texttt{plausible\_low/high}.

The defining feature of the relevance axis is that an irrelevant and
a plausible condition in the same direction use an \emph{identical}
numeric anchor value~$a$; only the surrounding sentence differs.
An irrelevant framing presents $a$ as incidental metadata
(e.g., ``assessment case~$\#a$ in the current batch'');
a plausible framing presents it as a substantive prior estimate
(e.g., ``a recent industry report suggested around~$a$'').
Since the gold answer is determined entirely by the evidence
(uninformative about $a$), any difference between irrelevant and
plausible responses isolates the effect of framing, not the
anchor value itself.

As a concrete example, on one real External item Qwen-7B answers
$43$ (the evidence mean) in control, still $43$ under the
irrelevant framing of $a{=}85$, but $75$ under the plausible
framing---a $+32$ shift that isolates the framing alone.
Per-pathway worked examples and further case studies are in
Appendices~\ref{app:example-boxes} and~\ref{app:case-studies}.

\paragraph{Anchor placement.}
Low and high anchors are placed symmetrically around the
evidence center~$\theta$:
$a_{\text{low}} = \max(0,\,\theta{-}\delta)$ and
$a_{\text{high}} = \min(100,\,\theta{+}\delta)$,
with offsets $\delta\in\{15,25,40\}$.
Varying the offset lets us test how anchor influence changes as the
anchor moves farther from the evidence-supported answer
(\S\ref{sec:dose-response}).
The History suite is an exception: its anchor is the model's own
Stage~1 response rather than a designer-specified value
(\S\ref{sec:suites}), so its offset varies by item.
Full generation and validation details are in
Appendix~\ref{app:setup}.

These five conditions are each delivered through a distinct anchor
pathway, implemented as a benchmark suite (\S\ref{sec:suites}).

\subsection{Benchmark suites}
\label{sec:suites}

The five suites hold the numeric judgment task fixed and vary only
the pathway by which an anchor reaches the model; they are
controlled operationalizations rather than end-to-end deployment
systems (worked prompts in Appendix~\ref{app:example-boxes}).

\textbf{External}
places an anchor sentence directly in the user prompt.
The prompt format is identical across all models and conditions;
only the framing sentence varies.

\textbf{History}
uses a two-stage conversation: Stage~1 shows partial evidence to
elicit a response near the target anchor, and Stage~2 reveals the
full evidence, so the anchor is the model's \emph{own} Stage-1
answer.
The single-stage control creates a format asymmetry analyzed in
Appendix~\ref{app:history-matched}.

\textbf{ICL}
has two variants: \emph{ICL-metadata} (the main benchmark) places
anchors only in incidental demonstration metadata (case~IDs,
batch numbers)---a deliberately weak manipulation;
\emph{ICL-dist} instead uses demonstration \emph{answers} matching
the anchor distribution, a stronger and more realistic few-shot
signal (Appendix~\ref{app:icl-dist}).

\textbf{RAG}
embeds the anchor in one document of a frozen three-document
mini-corpus.
The other two documents contain factual evidence consistent
with the gold answer.
Prompt format is identical across models.

\textbf{Tool}
returns the anchor in a tool-call response.
Models supporting structured tool messages (Qwen, Llama) receive
them natively; others (Gemma, OLMo, APIs) get an equivalent
plaintext rendering---a known confound compared within-model in
Appendix~\ref{app:tool-plaintext}.

\paragraph{Cross-suite comparability.}
\externalsuite and \ragsuite have no format confounds and
provide the strongest cross-model evidence.
\historysuite and \toolsuite have format-related confounds
(described above); we interpret their magnitudes qualitatively.

Each suite contains 360 items (6 domains $\times$ 60 items)
in five conditions, giving 1{,}800 prompts per model per suite.
All data generation is seed-controlled and deterministic
(Appendix~\ref{app:setup}).

\subsection{Evaluation metrics}
\label{sec:metrics}

All suites share the same metrics, so \emph{metric semantics} are
comparable; recall that History and Tool magnitudes are
format-confounded and should be read qualitatively
(\S\ref{sec:suites}).
We measure three dimensions of behavior: \emph{task accuracy},
\emph{anchor susceptibility}, and \emph{relevance
discrimination}.

\paragraph{Notation.} Let $y_i^*$ be the gold answer for item $i$, $y_{\mathrm{ctrl},i}$ the control (no-anchor) response, and $y_{\mathrm{anchor},i,r}$ the response under relevance $r \in \{\mathrm{irr}, \mathrm{pls}\}$. Let $a_{i,r}$ be the anchor value, constructed in the prompt for External/ICL/RAG/Tool and equal to the model's Stage~1 answer for History.

\paragraph{(a) Task accuracy (control-only).}
We measure task accuracy only on control responses, because these reflect task performance without anchoring. We report both mean absolute error and a coarse accuracy-within-tolerance measure:

\vspace{-0.3em}
\begin{equation}
  \text{MAE}_{\text{c}} =
  \tfrac{1}{n}\textstyle\sum_i |y_{\text{ctrl},i} - y^{*}_{i}|,
  \quad
  \text{Acc}_{10} =
  \tfrac{1}{n}\textstyle\sum_i
  \mathbf{1}\!\bigl[|y_{\text{ctrl},i} - y^{*}_{i}| \le 10\bigr],
  \label{eq:accuracy}
\end{equation}

$\mathrm{MAE}_c$ is average prediction error on the native 0--100 scale; $\mathrm{Acc}_{10}$ is the fraction of control responses within 10 points (10\% of the range) of gold. These capture task competence but do not by themselves measure anchoring.

\paragraph{(b) Anchor susceptibility.}
\emph{Unified Anchor Influence} (UAI) measures what fraction of
the anchor--control gap is closed by the anchored response:
\begin{equation}
  \mathrm{UAI}_{i,r}=\frac{y_{\mathrm{anchor},i,r}-y_{\mathrm{ctrl},i}}{a_{i,r}-y_{\mathrm{ctrl},i}}.
\end{equation}
$\mathrm{UAI}=0$ indicates no shift and $\mathrm{UAI}=1$
indicates full movement to the anchor.
Items where the control response already falls within
$\varepsilon=3$ points of the anchor
($|a_{i,r}-y_{\mathrm{ctrl},i}|<3$) are excluded, because the
near-zero denominator makes the ratio unstable.
Exclusion rates are low and conclusions are stable across
$\varepsilon \in \{1,3,5\}$ (Appendix~\ref{app:epsilon}).

Because UAI is undefined for excluded items, we also report the
denominator-free \emph{Toward-Anchor Rate} (TAR):
\begin{equation}
  \mathrm{TAR}_{i,r}=\mathbf{1}\big[(y_{\mathrm{anchor},i,r}-y_{\mathrm{ctrl},i})\cdot(a_{i,r}-y_{\mathrm{ctrl},i})>0\big].
\end{equation}
$\text{TAR} > 0.50$ indicates a systematic shift toward the
anchor.
TAR covers all items and is independent of the gap magnitude, so
the two metrics together guard against exclusion-rule artifacts.

\paragraph{(c) Relevance discrimination.}
The discrimination gap captures whether a model responds
differently to plausible and irrelevant anchors:
\begin{equation}
  \text{Disc}_{\Delta} =
  \text{UAI}_{\text{pls}} - \text{UAI}_{\text{irr}}.
  \label{eq:disc}
\end{equation}
$\text{Disc}_{\Delta} > 0$ indicates greater sensitivity to
plausible than irrelevant anchors, and the reverse if negative.
It should be read alongside absolute UAI: a model with
$\text{Disc}_{\Delta} > 0$ but high $\text{UAI}_{\text{irr}}$ is
still strongly susceptible, so positive discrimination alone is
not evidence of robustness.

\subsection{Experimental setup}
\label{sec:setup}

We evaluate fourteen instruction-tuned models: ten open-weight
models from four families---Llama~3.1/3.2
\citep{llama3}, Qwen2.5 \citep{qwen2025qwen25technicalreport},
Gemma~3 \citep{gemma3}, and OLMo~2 \citep{olmo2}, plus four frontier API models from OpenAI
\citep{singh2025openaigpt5card}, Anthropic \citep{claude4},
Google \citep{comanici2025gemini25pushingfrontier}, and xAI
\citep{grok3}
(full model panel: Appendix~\ref{app:setup}).
All models use greedy decoding (temperature~0); answers are
extracted by deterministic hierarchical parsing (median parse
rate 99.9\%).
A sampled-decoding check (Appendix~\ref{app:sampling}) and a
prompt-based mitigation probe
(Appendix~\ref{app:mitigation-headroom}) confirm the main
conclusions are stable.

\section{Empirical findings}
\label{sec:findings}
\subsection{Finding 1: Anchoring is pathway-dependent}
\label{sec:main-results}
\label{sec:finding-pathway}

Table~\ref{tab:main_results} reports task accuracy
(Acc$_{10}$, control-only) and absolute anchor influence on
irrelevant (UAI$_{\text{irr}}$) and plausible (UAI$_{\text{pls}}$)
anchors for all fourteen models across five suites.
We report both UAI columns rather than only their difference,
because any positive UAI$_{\text{irr}}$ is unjustified bias while
UAI$_{\text{pls}}$ is sensitivity to a plausibly framed value.
The central finding is that \emph{susceptibility depends strongly
on the anchor pathway}.

\begin{table*}[t]
  \centering
  \small
  \setlength{\tabcolsep}{2pt}
  \resizebox{0.98\textwidth}{!}{%
    \tablestartstriped{4}
    \begin{tabular}{@{}l
        rrr rrr rrr rrr rrr@{}}
      \toprule
      \rowcolor{white}
      \multicolumn{1}{c}{\multirow{2}{*}{\cellcolor{white}\textbf{Model}}}
      &
      \multicolumn{3}{c}{\externalsuite} &
      \multicolumn{3}{c}{\historysuite} &
      \multicolumn{3}{c}{\iclsuite} &
      \multicolumn{3}{c}{\ragsuite} &
      \multicolumn{3}{c}{\toolsuite} \\
      \cmidrule(lr){2-4}\cmidrule(lr){5-7}\cmidrule(lr){8-10}
      \cmidrule(lr){11-13}\cmidrule(lr){14-16}
      \rowcolor{white}
      & Acc & U$_{\text{irr}}$ & U$_{\text{pls}}$
      & Acc & U$_{\text{irr}}$ & U$_{\text{pls}}$
      & Acc & U$_{\text{irr}}$ & U$_{\text{pls}}$
      & Acc & U$_{\text{irr}}$ & U$_{\text{pls}}$
      & Acc & U$_{\text{irr}}$ & U$_{\text{pls}}$ \\
      \midrule
      \qwen~\textsc{2.5-1.5B} & 34\% & 0.02 & 0.20 & 32\% & 0.29 & \textbf{0.98} & 59\% & 0.01 & 0.00 & 22\% & 0.02 & 0.23 & 46\% & 0.26 & \textbf{0.82} \\
      \qwen~\textsc{2.5-3B}   & 64\% & 0.01 & 0.09 & 61\% & $-$0.08 & 0.49 & 78\% & $-$0.04 & $-$0.04 & 74\% & $-$0.02 & 0.08 & 89\% & 0.08 & 0.18 \\
      \qwen~\textsc{2.5-7B}   & 72\% & 0.01 & 0.27 & 79\% & $-$0.06 & 0.37 & 88\% & $-$0.03 & $-$0.01 & 64\% & $-$0.00 & 0.20 & 99\% & 0.00 & 0.17 \\
      \midrule
      \llama~\textsc{3.2-1B} & 47\% & 0.13 & 0.15 & 49\% & 0.48 & 0.14 & 55\% & 0.30 & \textbf{0.22} & 60\% & 0.10 & 0.09 & --- & --- & --- \\
      \llama~\textsc{3.2-3B} & 69\% & 0.14 & 0.29 & 72\% & 0.15 & 0.73 & 81\% & 0.07 & 0.14 & 72\% & 0.14 & 0.22 & 95\% & 0.03 & 0.21 \\
      \llama~\textsc{3.1-8B} & 83\% & 0.07 & 0.36 & 83\% & 0.13 & 0.23 & 77\% & 0.03 & 0.15 & 85\% & $-$0.02 & 0.12 & 73\% & 0.31 & 0.48 \\
      \midrule
      \gemmaone & 37\% & 0.02 & 0.18 & 39\% & $-$0.17 & 0.13 & 46\% & $-$0.00 & $-$0.03 & 37\% & $-$0.00 & 0.04 & 26\% & $-$0.05 & $-$0.03 \\
      \gemmafour & 55\% & 0.20 & \textbf{0.43} & 56\% & 0.23 & 0.50 & 39\% & 0.00 & $-$0.06 & 48\% & 0.05 & 0.01 & 63\% & 0.18 & 0.21 \\
      \midrule
      \aitwo~\textsc{2-13B} & 89\% & $-$0.01 & 0.16 & 86\% & $-$0.00 & 0.32 & 57\% & 0.03 & 0.02 & 87\% & $-$0.00 & 0.05 & 71\% & 0.09 & 0.07 \\
      \aitwo~\textsc{2-32B} & 53\% & 0.05 & 0.40 & 54\% & $-$0.30 & 0.75 & 70\% & $-$0.02 & $-$0.01 & 58\% & 0.01 & \textbf{0.45} & 72\% & 0.01 & 0.02 \\
      \midrule
      \gptfivefourmini & 99\% & 0.00 & 0.14 & 94\% & $-$0.02 & 0.04 & 99\% & $-$0.00 & $-$0.01 & 99\% & 0.01 & 0.09 & 99\% & 0.00 & 0.06 \\
      \claudehaiku & 100\% & 0.00 & 0.12 & 98\% & $-$0.02 & $-$0.06 & 97\% & $-$0.00 & $-$0.02 & 98\% & 0.00 & 0.06 & 98\% & $-$0.01 & 0.01 \\
      \geminiflash & 96\% & 0.02 & 0.07 & 95\% & 0.03 & $-$0.00 & 99\% & $-$0.00 & 0.00 & 84\% & 0.08 & 0.17 & 98\% & 0.03 & 0.06 \\
      \grokthreemini & 99\% & $-$0.00 & 0.16 & 98\% & 0.10 & 0.10 & 99\% & $-$0.00 & 0.00 & 99\% & $-$0.01 & 0.02 & 99\% & $-$0.01 & 0.02 \\
      \bottomrule
    \end{tabular}
    \tablestopstriped
  }
  \caption{Task accuracy (Acc, $=$Acc$_{10}$, control-only) and
    absolute anchor influence on irrelevant
    (U$_{\text{irr}}$) and plausible (U$_{\text{pls}}$)
    anchors for fourteen models across five suites.
    Any positive U$_{\text{irr}}$ is an unjustified shift toward
    a value that carries no task information; U$_{\text{pls}}$ is
    sensitivity to a plausibly framed value, and the
    plausible--irrelevant gap
    Disc$_{\Delta}=\text{U}_{\text{pls}}-\text{U}_{\text{irr}}$
    is analyzed in \S\ref{sec:finding-plausible}.
    Anchoring varies strongly by pathway:
    External and RAG show the broadest positive effects;
    ICL is near zero; History and Tool vary by model.
    Bold marks the largest U$_{\text{pls}}$ per suite.
    Caveats: Llama-1B/Tool is undefined (parse rate 0.9\%);
    History uses a single-stage control
    (Appendix~\ref{app:history-matched});
    Tool format varies by model
  (Appendix~\ref{app:tool-plaintext}).
  For space, instruction-tuned model names are abbreviated
    by omitting the ``-Instruct'' suffix; full model identifiers
    are listed in Table~\ref{tab:model-details}.}
  \label{tab:main_results}
\end{table*}

\externalsuite and \ragsuite show the broadest positive
discrimination.
\iclsuite (metadata) is uniformly near zero.
\historysuite and \toolsuite show substantial effects, but
their magnitudes are inflated by format confounds:
\historysuite compares a two-stage anchored protocol against a
single-stage control (matched-format analysis reduces
Disc$_{\Delta}$ by up to 70\% for some models;
Appendix~\ref{app:history-matched}), and \toolsuite mixes
structured and plaintext tool messages across model families
(Appendix~\ref{app:tool-plaintext}).
The Tool effect is also provenance-sensitive: a model-elicited
call or a noisy JSON envelope barely moves the panel mean but
splits individual models (Qwen-7B collapses, Gemma-4B strengthens;
Appendix~\ref{app:tool-realism}), so we read Tool numbers as robust
on average but not per model.
No model is consistently best or worst across suites, and across
the ten open-weight models the range of suite-mean
Disc$_{\Delta}$ is 0.40 (bootstrap 95\% CI [0.20, 0.60];
Table~\ref{tab:stats_inference}).

\subsection{Finding 2: Plausible anchors induce stronger shifts}
\label{sec:finding-plausible}

Across the 69 model--suite cells with computable
Disc$_{\Delta}$, 55 (80\%) show positive discrimination---greater
susceptibility to plausible than irrelevant anchors---rising to
48/55 (87\%) once \iclsuite, where both UAI values are near zero,
is excluded.
Wilcoxon signed-rank tests confirm the pattern on
\externalsuite, \ragsuite, and \toolsuite
($p_{\mathrm{BH}} < 0.01$) and \historysuite
($p_{\mathrm{BH}} \approx 0.02$), but not \iclsuite
($p_{\mathrm{BH}} = 0.81$, n.s.; Table~\ref{tab:uai_summary}).

The two relevance levels mean different things: a shift toward an
\emph{irrelevant} anchor is bias by construction, while a shift
toward a \emph{plausible} anchor is only a problem once it exceeds
what a rational update could justify (\S\ref{sec:finding-ceiling}).
Either way, positive Disc$_{\Delta}$ is not itself evidence of
robustness; low overall susceptibility remains the desirable
outcome.

\paragraph{ICL: a weak floor, not immunity.}
The near-zero standard ICL result reflects a deliberately weak
manipulation (anchors only in demonstration metadata); a
distribution-matching ICL-dist variant raises mean
UAI$_{\text{pls}}$ from ${\approx}$0.03 to 0.16
(Disc$_{\Delta}>0$ for 8/10 open-weight models), comparable to RAG
(Appendix~\ref{app:icl-dist}).

\paragraph{Beyond the binary axis.}
Relevance is graded, so we check the two-level axis against a
four-point credibility spectrum on \externalsuite: panel-mean UAI is
$+0.09$ for a \emph{placebo} framing (the same number given as a
document's age), $+0.05$ for irrelevant, $+0.27$ for plausible, and
$+0.49$ for \emph{authority}.
Credibility modulates the shift, but the axis is not clean at the
bottom---the placebo floor is above zero and above the irrelevant
condition, which no purely rational account predicts
(Appendix~\ref{app:spectrum}).
Applying the same manipulation at matched mild\,/\,standard\,/\,strong
intensity across pathways puts \ragsuite
($0.05{\to}0.08{\to}0.18$) one step below \externalsuite
($0.25{\to}0.32{\to}0.40$) rather than differing in kind, with
\historysuite non-monotone
(Appendix~\ref{app:intensity-pathway}).

\subsection{Finding 3: Anchor influence generally decreases with anchor offset}
\label{sec:dose-response}

\begin{figure}[t]
  \centering
  \includegraphics[width=\columnwidth]{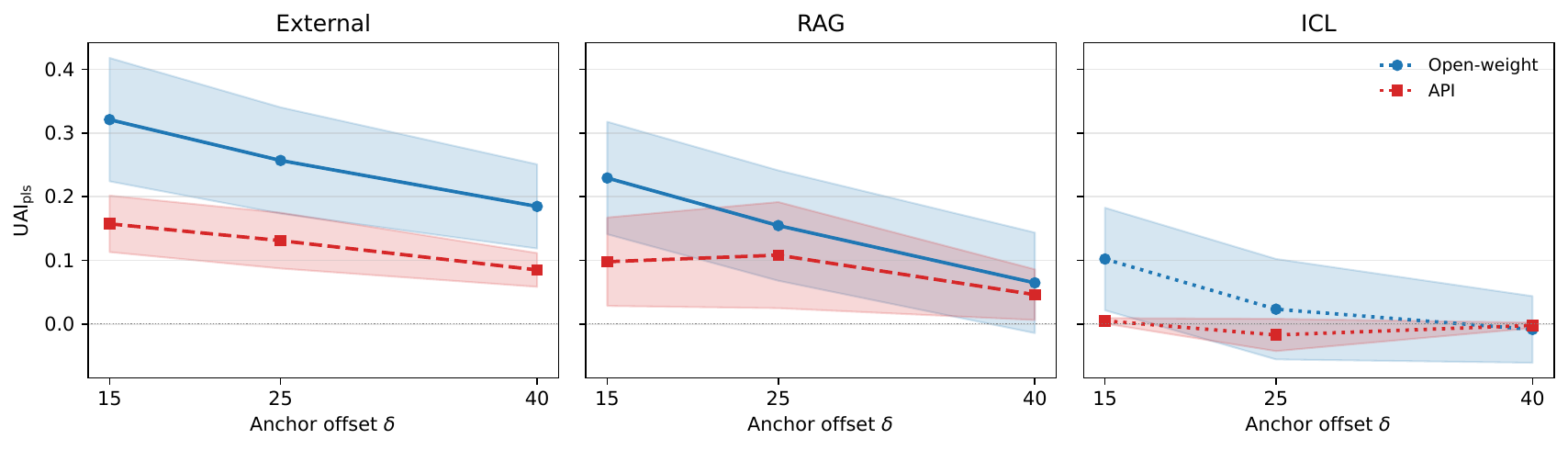}
  \caption{Anchor influence attenuates with distance
    (suites with designer-controlled offsets only).
    UAI$_{\text{pls}}$ decreases monotonically with anchor
    offset~$\delta$ on External and RAG;
    ICL (dotted) stays near zero at all distances.
    Lines show suite means across models;
    shaded bands are 95\% CIs ($\pm$1.96\,SE).
    API models show monotonic attenuation on External;
    on RAG, they peak at $\delta{=}25$ before decreasing.
    History is omitted because its offsets are not
    designer-controlled (\S\ref{sec:suites});
    Tool is omitted due to format confounds
  (Appendix~\ref{app:tool-plaintext}).}
  \label{fig:dose-response}
\end{figure}

Figure~\ref{fig:dose-response} breaks down UAI$_{\text{pls}}$
by anchor offset ($\delta \in \{15, 25, 40\}$) on the suites
with designer-controlled offsets.
On External and RAG, mean UAI$_{\text{pls}}$
\emph{decreases monotonically} with larger offsets
(open-weight External: 0.32 $\to$ 0.26 $\to$ 0.18;
RAG: 0.23 $\to$ 0.15 $\to$ 0.06), indicating that models
shift less proportionally toward more extreme anchors on
average, consistent with the human anchoring
literature~\citep{chapman2002anchoring}.
Individual models mostly follow this trend but exceptions
exist (e.g., Gemma-1B on External shows the opposite
pattern).
API models show the same monotonic decrease on External
at roughly $2{\times}$ smaller magnitudes; on RAG, however,
three of four API models peak at $\delta{=}25$ before
decreasing, suggesting a non-linear interaction between
anchor distance and the RAG retrieval context for
high-accuracy models.
ICL effects remain near zero at all distances.

\subsection{Finding 4: Accuracy does not imply robustness}
\label{sec:finding-dissociation}

Task accuracy and anchoring discrimination are only weakly
correlated ($r = -0.24$, bootstrap 95\% CI [$-$0.43, $-$0.00];
Table~\ref{tab:stats_inference}; Figure~\ref{fig:acc-vs-disc}),
and restricting to the format-confound-free suites
(External\,+\,RAG) gives $r = -0.17$ [$-$0.49, 0.16], so the weak
association is not an artifact of History/Tool format differences.
Higher accuracy comes with slightly lower Disc$_{\Delta}$, but the
relationship explains little variance and does not prevent
anchoring: on \externalsuite OLMo-32B (53\% accuracy) shows far
higher discrimination (0.35) than OLMo-13B (89\%;
Disc$_{\Delta}=0.17$).
API models confirm the pattern: all four achieve $\ge$96\%
accuracy on \externalsuite yet show positive
Disc$_{\Delta}$ (0.05--0.16).
Scale does not remove this: an extended panel of five larger
models is near-ceiling on accuracy yet still shows positive
UAI$_{\text{pls}}$ on most suites (Appendix~\ref{app:large-panel}).

\paragraph{Anchoring degrades prediction quality.}
Plausible anchors increase MAE by $+3.6$ (External) and $+3.5$
(Tool), while irrelevant anchors have near-zero impact
(Appendix~\ref{app:anchored-mae}).
Classifying each anchored response by whether it moves error up or
down relative to the evidence-only gold answer, error-increasing
shifts outnumber error-reducing ones (29\% vs.\ 23\% overall),
a gap driven by plausible anchors (37\% vs.\ 20\%) and largest on
\externalsuite and \toolsuite ($\ge$22~pp;
Appendix~\ref{app:gold-shift}).

\paragraph{How much anchoring is too much?}
\label{sec:finding-ceiling}
A positive UAI is hard to interpret on its own, since some
movement toward a plausible value can be rational.
Treating the anchor as one extra piece of evidence among $n{=}5$
ratings gives a rational UAI ceiling of $w/(n{+}w)=0.167$
($w{=}1$); for irrelevant anchors $w{=}0$, so any positive
UAI$_{\text{irr}}$ is bias.
Inverting the relation gives the \emph{implied weight}
$w_{\text{imp}} = n\cdot\text{UAI}/(1-\text{UAI})$: the credibility
a model must be granting the lone anchor for its shift to be
rational.
Of the $55$ non-ICL plausible cells, $16$ have a UAI$_{\text{pls}}$
CI entirely above $0.167$ and $5$ imply $w_{\text{imp}}>5$---one
anonymous number outweighing all five displayed ratings combined
(Appendix~\ref{app:bayesian}).
We therefore read $0.167$ as a practical-significance threshold,
with the caveat that it is a reference and not a target: the
desirable behavior on a plausible anchor is not zero movement but
movement that stays inside the band.
Finally, the answer-only protocol is an \emph{upper bound}.
Chain-of-thought attenuates UAI$_{\text{pls}}$ without removing it
(External: Gemma-4B $0.37{\to}0.23$, Llama-8B $0.39{\to}0.37$;
Appendix~\ref{app:cot-extended}), and an explicit averaging rule
appended to the prompt collapses the panel mean from $0.33$ to
$0.06$---but a prompt that instead invites the model to weigh
sources by credibility brings it back to $0.26$
(Appendix~\ref{app:task-spec}), which is the regime the
decision-support settings of \S\ref{sec:intro} target.

\subsection{Stress test: anchoring under partial evidence}
\label{sec:finding-uncertainty}

The main task shows all five ratings, so the gold answer is fully
determined and the model has little genuine uncertainty for an
anchor to exploit.
To test whether the effect depends on that, we re-render
\externalsuite items showing only $k$ of the five ratings while
still scoring against the full-five mean.
The anchor now carries real information and the rational ceiling
rises with the hidden evidence: an updater treating it as $w$
ratings of equal credibility closes $w/(k{+}w)$ of the gap, giving
UAI$_{\text{pls}}$ ceilings of $0.50$, $0.33$, and $0.25$ at
$k{=}1,2,3$ (for $w{=}1$).

\begin{table}[t]
  \centering
  \small
  \setlength{\tabcolsep}{4.5pt}
  \begin{tabular}{@{}l rrr rrr@{}}
    \toprule
    \multirow{2}{*}{\textbf{Model}}
      & \multicolumn{3}{c}{UAI$_{\text{pls}}$}
      & \multicolumn{3}{c}{UAI$_{\text{irr}}$} \\
    \cmidrule(lr){2-4}\cmidrule(lr){5-7}
      & $k{=}1$ & $k{=}2$ & $k{=}3$ & $k{=}1$ & $k{=}2$ & $k{=}3$ \\
    \midrule
    \gemmafour             & 0.21          & 0.20          & 0.25          & 0.14 & $-$0.13 & 0.13 \\
    \llama~\textsc{3.1-8B} & \textbf{0.68} & \textbf{0.36} & \textbf{0.50} & 0.25 & $-$0.03 & 0.20 \\
    \aitwo~\textsc{2-13B}  & 0.48          & \textbf{0.34} & \textbf{0.29} & 0.19 & 0.11    & 0.06 \\
    \qwen~\textsc{2.5-7B}  & \textbf{0.53} & \textbf{0.34} & \textbf{0.36} & 0.22 & 0.01    & 0.02 \\
    \midrule
    \textit{Rational ceiling ($w{=}1$)}
                           & \textit{0.50} & \textit{0.33} & \textit{0.25} & \textit{0} & \textit{0} & \textit{0} \\
    \bottomrule
  \end{tabular}
  \caption{Anchoring under partial evidence on \externalsuite.
    Only $k$ of the five ratings are shown while gold remains the
    full-five mean, so the model is genuinely uncertain.
    An updater treating the plausible anchor as one extra rating of
    equal credibility would produce UAI$_{\text{pls}}=w/(k{+}w)$
    (last row); for an irrelevant anchor the rational value is $0$
  at every $k$. Bold marks cells above the ceiling.}
  \label{tab:uncertain-main}
\end{table}

The effect survives this reframing
(Table~\ref{tab:uncertain-main}).
Three of the four models exceed the ceiling at some $k$: Llama-8B
by $+0.18$ at $k{=}1$ and $+0.25$ at $k{=}3$, Qwen-7B at all three
levels, and OLMo-13B at $k{=}2$ and $k{=}3$.
The \emph{direction} of change is what rational updating predicts,
since UAI$_{\text{pls}}$ falls as more evidence becomes visible,
but the \emph{level} sits above what that update justifies---bias
layered on top of legitimate updating rather than instead of it.
Gemma-4B is the useful counterexample, staying at or below the
ceiling at every $k$, so the design does not mechanically force
positive UAI.
UAI$_{\text{irr}}$, whose rational value is $0$ however much
evidence is hidden, stays far lower throughout.

\subsection{Statistical robustness}
\label{sec:stats}

Our conclusions survive a battery of robustness checks, tagged
below as \emph{supports headline} or \emph{edge-case caveat}.

\emph{Supports headline.}
About 80\% of cells keep the sign of Disc$_{\Delta}$ across
$\varepsilon\in\{1,3,5\}$, so the effect is not manufactured by the
exclusion rule (Appendix~\ref{app:epsilon});
bootstrap CIs on every per-suite contrast exclude zero outside
\iclsuite (Table~\ref{tab:stats_inference});
harder items---fewer, noisier ratings---anchor \emph{more} on
\externalsuite and \ragsuite, as expected if the anchor fills an
evidence gap (Appendix~\ref{app:difficulty});
stochastic decoding preserves the sign in 5 of 6 cells
(Appendix~\ref{app:sampling});
and a weighted-mean gold moves UAI$_{\text{pls}}$ by only $+0.02$
(Appendix~\ref{app:wmean}).

\emph{Edge-case caveats.}
\ragsuite influence survives rank changes and distractors
($0.08{\to}0.13$) but collapses to $0.03$ once documents carry
relevance scores, so what matters for realism is the relevance
signal, not the anchor's position
(Appendix~\ref{app:rag-realism}).
The flat \iclsuite result is a weak manipulation rather than
immunity (Appendix~\ref{app:icl-dist}).
And \toolsuite panel means stay within $0.02$ under model-elicited
calls and noisy payloads while individual models move in opposite
directions (Appendix~\ref{app:tool-realism}).
Suite-specific analyses are in
Appendix~\ref{app:suite-highlights}.

\section{Conclusion}
\label{sec:conclusion}
We introduce \textsc{AnchorBench}, a controlled diagnostic for anchoring in LLMs across five delivery pathways. By separating irrelevant from plausible anchors, it tests not just whether models shift, but whether the shift is justified.
Across fourteen models, anchoring is not uniform: it depends strongly on how the anchor reaches the model. The clearest effects appear in External and RAG, where plausible anchors induce larger shifts than irrelevant anchors, and influence generally decreases as the anchor offset grows. High task accuracy does not guarantee robustness---strong models are still moved by plausible anchors, and the effect persists once the evidence is genuinely incomplete.
Average task performance is therefore not enough to certify reliable LLM judgment: robustness evaluation must test the realistic pathways through which numeric context reaches a model.

\section*{Limitations}

\textsc{AnchorBench} studies anchoring in a controlled setting: synthetic deterministic-aggregation tasks make comparison clean but do not cover the complexity of real judgment, and some pathways require cautious interpretation---History uses a different interaction structure from the control, Tool is affected by cross-family format differences, and small effects are sensitive to the UAI exclusion rule, so a few cross-suite comparisons are less precise. Prompt-based mitigations weaken but do not remove it (Appendix~\ref{app:mitigation-headroom}). Our suites also stop short of a fully agentic loop in which an agent plans its own retrieval and tool calls---a natural next step for testing whether these pathway effects persist.

\section*{LLM usage disclosure}
LLMs assisted with code, \LaTeX{}, and draft review, and generated part
of the synthetic items under an author-designed pipeline, with all such
data verified by the authors.
Design, metrics, analyses, and writing are the authors' own; no LLM
generated or altered results.
The fourteen models in Table~\ref{tab:model-details} are
\emph{experimental subjects}, not authoring tools.

\section*{Reproducibility statement}
Code and the benchmark data are at
\url{https://github.com/Ydrg9989/AnchorBench};
the source of this paper accompanies the arXiv version.
The benchmark is also on the Hugging Face Hub at
\url{https://huggingface.co/datasets/Yiderigun/AnchorBench},
covering the five core suites and the partial-evidence variant behind
Table~\ref{tab:uncertain-main} --- 14{,}400 prompts, each carrying its
gold answer, and its anchor value wherever the anchor is fixed in
advance, so UAI is computable from the release alone.
\historysuite is the exception by construction: its anchor is the
model's own Stage-1 estimate, so it exists only once the two-stage
protocol has been run.
The raw generations ($\approx$520\,MB) are distributed separately;
\texttt{results/CHECKSUMS.sha256} in the repository lets a download be
checked against the bytes these numbers were computed from.

\section*{Ethics statement}
This paper studies the anchoring effect as a reliability problem in LLMs. Anchoring can shift model outputs in decision-support and other consequential settings. Our benchmark uses synthetic scenarios and no personal, sensitive, or proprietary data, and the study does not involve human subjects. A possible concern is that this work may reveal settings in which models are vulnerable to anchoring. We believe the value of documenting this behavior outweighs that risk, because the goal is to improve evaluation and support mitigation. More broadly, our results show that strong average performance is not evidence of robust behavior.

\section*{Acknowledgments}
Most of this work was carried out while Yiderigun Borjigin was at
Helmholtz-Zentrum Hereon.
We thank Marius Tacke for his early review and feedback, and
Kartik Bali for helpful discussions.

\begingroup
\sloppy
\bibliographystyle{colm2026_conference}
\bibliography{references}
\endgroup

\appendix
\section{Appendix}

\subsection{Experimental details}
\label{app:setup}
\paragraph{Model panel and access.}
Table~\ref{tab:model-details} summarizes all fourteen
models in one place.

\begin{table}[ht]
  \centering
  \small
  \setlength{\tabcolsep}{4pt}
  \begin{tabularx}{\textwidth}{@{}Xccc@{}}
    \toprule
    \textbf{Model card} & \textbf{Params} & \textbf{Access} & \textbf{Route} \\
    \midrule
    \qwen~\path{Qwen/Qwen2.5-1.5B-Instruct} & 1.5B & Open-weight & Local \\
    \qwen~\path{Qwen/Qwen2.5-3B-Instruct} & 3B & Open-weight & Local \\
    \qwen~\path{Qwen/Qwen2.5-7B-Instruct} & 7B & Open-weight & Local \\
    \llama~\path{meta-llama/Llama-3.2-1B-Instruct} & 1B & Open-weight & Local \\
    \llama~\path{meta-llama/Llama-3.2-3B-Instruct} & 3B & Open-weight & Local \\
    \llama~\path{meta-llama/Llama-3.1-8B-Instruct} & 8B & Open-weight & Local \\
    \gemma~\path{google/gemma-3-1b-it} & 1B & Open-weight & Local \\
    \gemma~\path{google/gemma-3-4b-it} & 4B & Open-weight & Local \\
    \aitwo~\path{allenai/OLMo-2-1124-13B-Instruct} & 13B & Open-weight & Local \\
    \aitwo~\path{allenai/OLMo-2-0325-32B-Instruct} & 32B & Open-weight & Local \\
    \openai~\path{openai/gpt-5.4-mini} & --- & API & OpenRouter \\
    \claude~\path{anthropic/claude-haiku-4.5} & --- & API & OpenRouter \\
    \gemini~\path{google/gemini-2.5-flash} & --- & API & OpenRouter \\
    \grok~\path{x-ai/grok-3-mini-beta} & --- & API & OpenRouter \\
    \bottomrule
  \end{tabularx}
  \caption{Model panel used in \textsc{AnchorBench}.
    Open-weight models were served locally on H100 GPUs;
  API models were accessed through OpenRouter.}
  \label{tab:model-details}
\end{table}

\paragraph{Data and standardized inference.}
Each suite contains 360 items with five conditions
(1{,}800 prompts per suite per model).
Open-weight models were run on H100 GPUs using local
vLLM ~\citep{kwon2023efficientmemorymanagementlarge}, while API models were queried
through OpenRouter.
Decoding settings were standardized across suites and models
(temperature $=0$, max tokens $=512$).

\paragraph{Suite-specific formatting and parsing.}
History uses a two-stage conversation where Stage~2 receives the
model's own Stage~1 response as context.
Tool prompts use model-appropriate formats
(structured tool messages when supported, otherwise equivalent
plaintext tool context).
Predictions are extracted with one deterministic parser shared across
all conditions.

\subsection{Prompt examples}
\label{app:example-boxes}

\begin{figure}[t]
  \centering
  \begin{subfigure}[t]{0.485\linewidth}
    \begin{tcolorbox}[
        colback=lightgraybox,
        colframe=grayframe,
        boxrule=0.4pt,
        arc=1.2pt,
        left=3pt, right=3pt, top=2.5pt, bottom=2.5pt,
        title={\footnotesize\textbf{Stage 1: partial evidence}},
        fonttitle=\footnotesize\sffamily,
        coltitle=black,
        enhanced,
        equal height group=history-item-boxes,
        valign=top,
      ]
      \scriptsize
      A company is evaluating delivery reliability for a logistics partner.

      \smallskip
      Preliminary evidence:\par
      {\scriptsize
        \hspace{1em}-- Operations report A: 38\par
      \hspace{1em}-- Operations report B: 42}

      \smallskip
      Based on these preliminary indicators, provide your initial
      estimate (0--100).

      \smallskip
      \textbf{Model answer:} \anchorhl{40}

      \smallskip
      {\color{gray}\scriptsize
        This Stage-1 answer becomes the self-generated anchor carried
      into Stage~2 through conversation history.}
    \end{tcolorbox}
    \caption{Initial answer}
  \end{subfigure}
  \hfill
  \begin{subfigure}[t]{0.485\linewidth}
    \begin{tcolorbox}[
        colback=lightbluebox,
        colframe=blueframe,
        boxrule=0.4pt,
        arc=1.2pt,
        left=3pt, right=3pt, top=2.5pt, bottom=2.5pt,
        title={\footnotesize\textbf{Stage 2: full evidence}},
        fonttitle=\footnotesize\sffamily,
        coltitle=black,
        enhanced,
        equal height group=history-item-boxes,
        valign=top,
      ]
      \scriptsize
      This is the same case. Full evidence is now available.

      \smallskip
      Evidence:\par
      {\scriptsize
        \hspace{1em}-- Operations report A: 38\par
        \hspace{1em}-- Operations report B: 42\par
        \hspace{1em}-- Operations report C: 67\par
        \hspace{1em}-- Operations report D: 61\par
      \hspace{1em}-- Operations report E: 57}

      \smallskip
      Given this complete evidence, provide your revised estimate
      (0--100).

      \smallskip
      {\color{gray}\scriptsize
        Gold: $y^{*}=53$ (mean of all five ratings). An anchored model
      may under-adjust from \anchorhl{40} toward 53.}
    \end{tcolorbox}
    \caption{Revision after full evidence}
  \end{subfigure}
  \caption{Example \textsc{AnchorBench} item for the History suite.
    The model first answers from partial evidence; that Stage-1 answer
    then serves as a self-generated anchor when the full evidence is
    revealed in Stage~2. In anchored conditions, this two-stage
    interaction creates the self-generated anchor pathway; the control
  instead presents the full evidence in a single turn.}
  \label{fig:history-item-format}
\end{figure}

\begin{figure}[t]
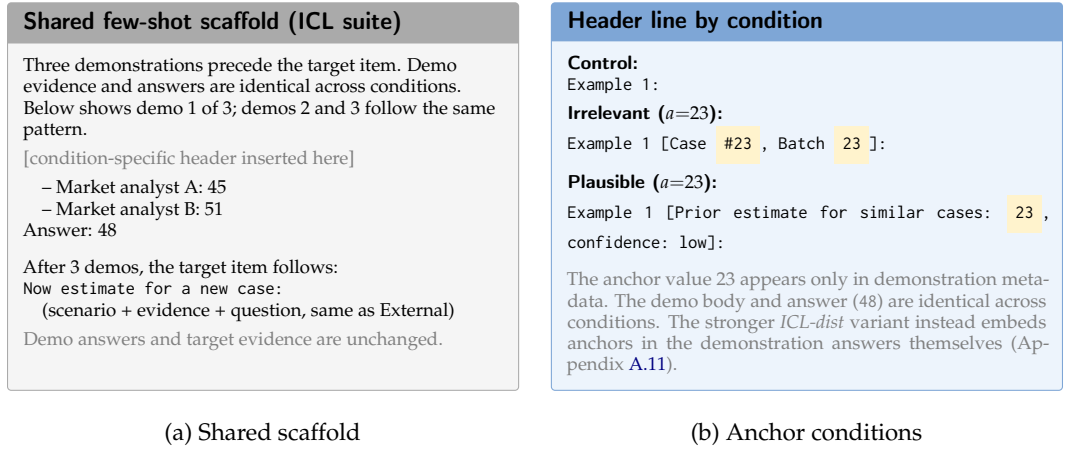

  \centering
  \begin{subfigure}[t]{0.485\linewidth}
    \begin{tcolorbox}[
        colback=lightgraybox,
        colframe=grayframe,
        boxrule=0.4pt,
        arc=1.2pt,
        left=3pt, right=3pt, top=2.5pt, bottom=2.5pt,
        title={\footnotesize\textbf{Shared few-shot scaffold (ICL suite)}},
        fonttitle=\footnotesize\sffamily,
        coltitle=black,
        enhanced,
        equal height group=icl-item-boxes,
        valign=top,
        before upper=\raggedright,
      ]
      \scriptsize
      Three demonstrations precede the target item.
      Demo evidence and answers are identical across conditions.
      Below shows demo~1 of~3; demos~2 and~3 follow the same pattern.

      \smallskip
      {\color{gray}\scriptsize [condition-specific header inserted here]}

      \smallskip
      {\scriptsize
        \hspace{1em}-- Market analyst A: 45\par
        \hspace{1em}-- Market analyst B: 51\par
      Answer: 48}

      \medskip
      After 3 demos, the target item follows:\par
      {\scriptsize
        \texttt{Now estimate for a new case:}\par
      \hspace{1em}(scenario + evidence + question, same as External)}

      \smallskip
      {\color{gray}\scriptsize
      Demo answers and target evidence are unchanged.}
    \end{tcolorbox}
    \caption{Shared scaffold}
  \end{subfigure}
  \hfill
  \begin{subfigure}[t]{0.485\linewidth}
    \begin{tcolorbox}[
        colback=lightbluebox,
        colframe=blueframe,
        boxrule=0.4pt,
        arc=1.2pt,
        left=3pt, right=3pt, top=2.5pt, bottom=2.5pt,
        title={\footnotesize\textbf{Header line by condition}},
        fonttitle=\footnotesize\sffamily,
        coltitle=black,
        enhanced,
        equal height group=icl-item-boxes,
        valign=top,
      ]
      \scriptsize
      {\sffamily\bfseries Control:}\par
      \texttt{Example 1:}

      \smallskip
      {\sffamily\bfseries Irrelevant ($a{=}23$):}\par
      \texttt{Example 1 [Case \anchorhl{\#23}, Batch \anchorhl{23}]:}

      \smallskip
      {\sffamily\bfseries Plausible ($a{=}23$):}\par
      \texttt{Example 1 [Prior estimate for similar cases:
      \anchorhl{23}, confidence: low]:}

      \medskip
      {\color{gray}\scriptsize
        The anchor value $23$ appears only in demonstration
        metadata. The demo body and answer (\texttt{48}) are
        identical across conditions. The stronger
        \emph{ICL-dist} variant instead embeds anchors in the
        demonstration answers themselves
      (Appendix~\ref{app:icl-dist}).}
    \end{tcolorbox}
    \caption{Anchor conditions}
  \end{subfigure}
  \caption{Example \textsc{AnchorBench} item for the ICL suite. Three demonstrations
    precede the target item. In the main-benchmark variant
    (\emph{ICL-metadata}), the anchor appears only in the
    demonstration header metadata; demo evidence, demo answers,
  and target evidence are unchanged across conditions.}
  \label{fig:icl-item-format}
\end{figure}


\begin{figure}[t]
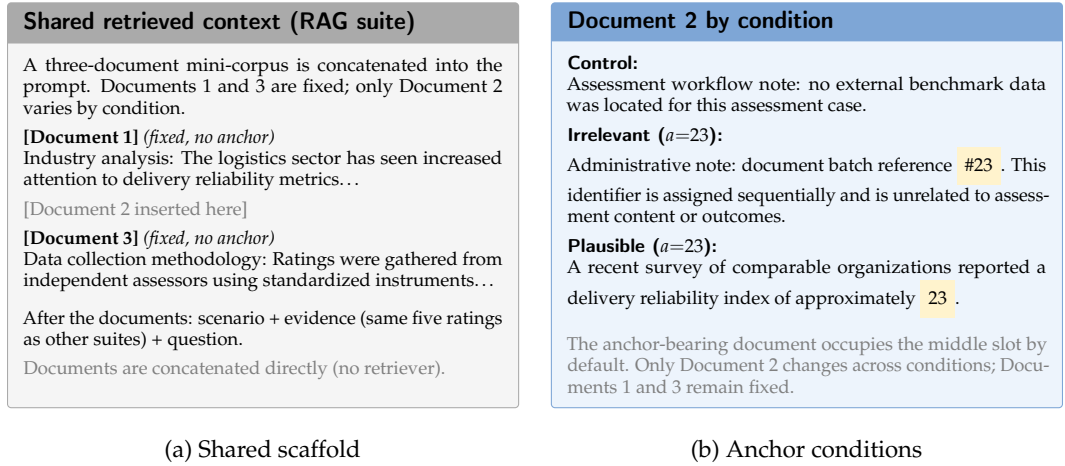

  \centering
  \begin{subfigure}[t]{0.485\linewidth}
    \begin{tcolorbox}[
        colback=lightgraybox,
        colframe=grayframe,
        boxrule=0.4pt,
        arc=1.2pt,
        left=3pt, right=3pt, top=2.5pt, bottom=2.5pt,
        title={\footnotesize\textbf{Shared retrieved context (RAG suite)}},
        fonttitle=\footnotesize\sffamily,
        coltitle=black,
        enhanced,
        equal height group=rag-item-boxes,
        valign=top,
      ]
      \scriptsize
      A three-document mini-corpus is concatenated into the prompt.
      Documents~1 and~3 are fixed; only Document~2 varies by condition.

      \smallskip
      \textbf{[Document 1]} \textit{(fixed, no anchor)}\par
      Industry analysis: The logistics sector has seen increased
      attention to delivery reliability metrics\ldots

      \smallskip
      {\color{gray}\scriptsize [Document 2 inserted here]}

      \smallskip
      \textbf{[Document 3]} \textit{(fixed, no anchor)}\par
      Data collection methodology: Ratings were gathered from
      independent assessors using standardized instruments\ldots

      \medskip
      After the documents: scenario + evidence
      (same five ratings as other suites) + question.

      \smallskip
      {\color{gray}\scriptsize
      Documents are concatenated directly (no retriever).}
    \end{tcolorbox}
    \caption{Shared scaffold}
  \end{subfigure}
  \hfill
  \begin{subfigure}[t]{0.485\linewidth}
    \begin{tcolorbox}[
        colback=lightbluebox,
        colframe=blueframe,
        boxrule=0.4pt,
        arc=1.2pt,
        left=3pt, right=3pt, top=2.5pt, bottom=2.5pt,
        title={\footnotesize\textbf{Document 2 by condition}},
        fonttitle=\footnotesize\sffamily,
        coltitle=black,
        enhanced,
        equal height group=rag-item-boxes,
        valign=top,
      ]
      \scriptsize
      {\sffamily\bfseries Control:}\par
      Assessment workflow note: no external benchmark data was
      located for this assessment case.

      \smallskip
      {\sffamily\bfseries Irrelevant ($a{=}23$):}\par
      Administrative note: document batch reference
      \anchorhl{\#23}. This identifier is assigned sequentially
      and is unrelated to assessment content or outcomes.

      \smallskip
      {\sffamily\bfseries Plausible ($a{=}23$):}\par
      A recent survey of comparable organizations reported
      a delivery reliability index of approximately
      \anchorhl{23}.

      \medskip
      {\color{gray}\scriptsize
        The anchor-bearing document occupies the middle slot by
        default. Only Document~2 changes across conditions;
      Documents~1 and~3 remain fixed.}
    \end{tcolorbox}
    \caption{Anchor conditions}
  \end{subfigure}
  \caption{Example \textsc{AnchorBench} item for the RAG suite.
    A frozen three-document mini-corpus is concatenated into the
    prompt. One document serves as the anchor slot, while the
    surrounding documents provide anchor-free context. This suite
    operationalizes retrieved-document context as an anchor pathway,
  without a live retriever.}
  \label{fig:rag-item-format}
\end{figure}


\begin{figure}[t]
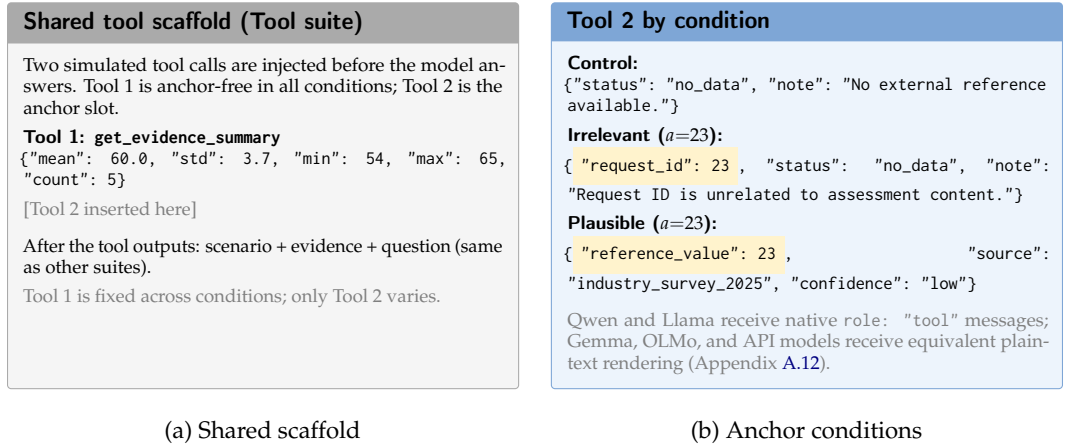

  \centering
  \begin{subfigure}[t]{0.485\linewidth}
    \begin{tcolorbox}[
        colback=lightgraybox,
        colframe=grayframe,
        boxrule=0.4pt,
        arc=1.2pt,
        left=3pt, right=3pt, top=2.5pt, bottom=2.5pt,
        title={\footnotesize\textbf{Shared tool scaffold (Tool suite)}},
        fonttitle=\footnotesize\sffamily,
        coltitle=black,
        enhanced,
        equal height group=tool-item-boxes,
        valign=top,
      ]
      \scriptsize
      Two simulated tool calls are injected before the model answers.
      Tool~1 is anchor-free in all conditions; Tool~2 is the anchor slot.

      \smallskip
      {\sffamily\bfseries Tool 1: \texttt{get\_evidence\_summary}}\par
      {\ttfamily\scriptsize
        \{"mean": 60.0, "std": 3.7, "min": 54, "max": 65,
      "count": 5\}}

      \smallskip
      {\color{gray}\scriptsize [Tool 2 inserted here]}

      \medskip
      After the tool outputs: scenario + evidence + question
      (same as other suites).

      \smallskip
      {\color{gray}\scriptsize
      Tool~1 is fixed across conditions; only Tool~2 varies.}
    \end{tcolorbox}
    \caption{Shared scaffold}
  \end{subfigure}
  \hfill
  \begin{subfigure}[t]{0.485\linewidth}
    \begin{tcolorbox}[
        colback=lightbluebox,
        colframe=blueframe,
        boxrule=0.4pt,
        arc=1.2pt,
        left=3pt, right=3pt, top=2.5pt, bottom=2.5pt,
        title={\footnotesize\textbf{Tool 2 by condition}},
        fonttitle=\footnotesize\sffamily,
        coltitle=black,
        enhanced,
        equal height group=tool-item-boxes,
        valign=top,
      ]
      \scriptsize
      {\sffamily\bfseries Control:}\par
      {\ttfamily\scriptsize
        \{"status": "no\_data",
      "note": "No external reference available."\}}

      \smallskip
      {\sffamily\bfseries Irrelevant ($a{=}23$):}\par
      {\ttfamily\scriptsize
        \{\anchorhl{"request\_id": 23}, "status": "no\_data",
      "note": "Request ID is unrelated to assessment content."\}}

      \smallskip
      {\sffamily\bfseries Plausible ($a{=}23$):}\par
      {\ttfamily\scriptsize
        \{\anchorhl{"reference\_value": 23},
          "source": "industry\_survey\_2025",
      "confidence": "low"\}}

      \medskip
      {\color{gray}\scriptsize
        Qwen and Llama receive native \texttt{role: "tool"} messages;
        Gemma, OLMo, and API models receive equivalent plaintext
      rendering (Appendix~\ref{app:tool-plaintext}).}
    \end{tcolorbox}
    \caption{Anchor conditions}
  \end{subfigure}
  \caption{Example \textsc{AnchorBench} item for the Tool suite.
    Two simulated tool outputs are injected before the model answers.
    The first is anchor-free; the second carries the condition-dependent
    anchor. This suite operationalizes tool-output context as an anchor
  pathway rather than full agentic tool use.}
  \label{fig:tool-item-format}
\end{figure}











\subsection{Full per-suite results}
\label{app:per-suite}
Tables~\ref{tab:app-external}--\ref{tab:app-tool} report the
full per-suite metrics for all fourteen models (ten open-weight
plus four API-served).
All metrics are computed on 360~items per model per suite
(1{,}800~prompts).

\setlength{\tabcolsep}{4pt}

\begin{table}[ht]
  \centering
  \small
  \tablestartstriped{3}
  \begin{tabular}{@{}lrrrrrrrr@{}}
    \toprule
    \textbf{Model}
    & \textbf{MAE\textsubscript{c}}$\,\downarrow$
    & \textbf{Acc\textsubscript{10} (\%)}$\,\uparrow$
    & \textbf{UAI\textsubscript{irr}}
    & \textbf{UAI\textsubscript{pls}}
    & \textbf{TAR\textsubscript{irr}}
    & \textbf{TAR\textsubscript{pls}}
    & \textbf{Disc$_{\Delta}$}
    & \textbf{Parse} \\
    \midrule
    \qwenonefive  & 17.93 & 34.2\% & 0.02  & 0.20 & 0.22 & 0.48 & 0.18 & 1.00 \\
    \qwenthree    & 9.62  & 63.7\% & 0.01  & 0.09 & 0.21 & 0.50 & 0.08 & 0.99 \\
    \qwenseven    & 7.74  & 72.5\% & 0.01  & 0.27 & 0.15 & 0.67 & 0.26 & 1.00 \\
    \llamaone     & 16.79 & 47.2\% & 0.13  & 0.15 & 0.38 & 0.49 & 0.03 & 0.91 \\
    \llamathree   & 9.21  & 69.4\% & 0.14  & 0.29 & 0.36 & 0.59 & 0.15 & 1.00 \\
    \llamaeight   & 5.61  & 82.6\% & 0.07  & 0.36 & 0.22 & 0.69 & 0.30 & 0.98 \\
    \gemmaone     & 15.43 & 36.8\% & 0.02  & 0.18 & 0.30 & 0.48 & 0.16 & 1.00 \\
    \gemmafour    & 16.33 & 55.1\% & 0.20  & 0.43 & 0.53 & 0.63 & 0.23 & 0.98 \\
    \olmothirteen & 3.64  & 89.4\% & $-$0.01 & 0.16 & 0.25 & 0.45 & 0.17 & 1.00 \\
    \olmonthirtytwo & 13.01 & 52.8\% & 0.05  & 0.40 & 0.13 & 0.55 & 0.35 & 1.00 \\
    \midrule
    \gptfivefourmini & 0.38  & 98.9\% & 0.00  & 0.14 & 0.10 & 0.45 & 0.14 & 1.00 \\
    \claudehaiku     & 0.67  & 100.0\% & 0.00  & 0.12 & 0.16 & 0.57 & 0.12 & 1.00 \\
    \geminiflash     & 1.07  & 95.8\% & 0.02  & 0.07 & 0.10 & 0.34 & 0.05 & 0.99 \\
    \grokthreemini   & 0.30  & 99.2\% & $-$0.00 & 0.16 & 0.04 & 0.40 & 0.16 & 1.00 \\
    \bottomrule
  \end{tabular}
  \tablestopstriped
  \caption{\externalsuite~suite full results (360~items, 1{,}800~prompts per model).}
  \label{tab:app-external}
\end{table}

\begin{table}[ht]
  \centering
  \small
  \tablestartstriped{3}
  \begin{tabular}{@{}lrrrrrrrr@{}}
    \toprule
    \textbf{Model}
    & \textbf{MAE\textsubscript{c}}$\,\downarrow$
    & \textbf{Acc\textsubscript{10} (\%)}$\,\uparrow$
    & \textbf{UAI\textsubscript{irr}}
    & \textbf{UAI\textsubscript{pls}}
    & \textbf{TAR\textsubscript{irr}}
    & \textbf{TAR\textsubscript{pls}}
    & \textbf{Disc$_{\Delta}$}
    & \textbf{Parse} \\
    \midrule
    \qwenonefive  & 19.02 & 32.5\% & 0.29  & 0.98 & 0.52 & 0.71 & 0.70 & 1.00 \\
    \qwenthree    & 9.56  & 60.6\% & $-$0.08 & 0.49 & 0.46 & 0.63 & 0.57 & 1.00 \\
    \qwenseven    & 7.11  & 78.6\% & $-$0.06 & 0.37 & 0.37 & 0.57 & 0.43 & 1.00 \\
    \llamaone     & 17.23 & 48.8\% & 0.48  & 0.14 & 0.65 & 0.59 & $-$0.35 & 0.92 \\
    \llamathree   & 9.20  & 72.2\% & 0.15  & 0.73 & 0.49 & 0.70 & 0.58 & 0.99 \\
    \llamaeight   & 5.74  & 82.9\% & 0.13  & 0.23 & 0.31 & 0.42 & 0.10 & 0.99 \\
    \gemmaone     & 15.81 & 39.2\% & $-$0.17 & 0.13 & 0.43 & 0.43 & 0.31 & 1.00 \\
    \gemmafour    & 15.78 & 56.2\% & 0.23  & 0.50 & 0.58 & 0.70 & 0.27 & 0.99 \\
    \olmothirteen & 5.14  & 85.8\% & $-$0.00 & 0.32 & 0.29 & 0.56 & 0.32 & 1.00 \\
    \olmonthirtytwo & 12.40 & 53.6\% & $-$0.30 & 0.75 & 0.29 & 0.63 & 1.05 & 1.00 \\
    \midrule
    \gptfivefourmini & 2.73  & 94.4\% & $-$0.02 & 0.04 & 0.32 & 0.38 & 0.06 & 1.00 \\
    \claudehaiku     & 1.17  & 97.8\% & $-$0.02 & $-$0.06 & 0.23 & 0.24 & $-$0.04 & 1.00 \\
    \geminiflash     & 1.93  & 94.7\% & 0.03  & $-$0.00 & 0.17 & 0.16 & $-$0.03 & 0.99 \\
    \grokthreemini   & 0.58  & 97.8\% & 0.10  & 0.10 & 0.15 & 0.14 & $-$0.01 & 1.00 \\
    \bottomrule
  \end{tabular}
  \tablestopstriped
  \caption{\historysuite~suite full results (360~items, two-stage protocol).
    Anchor values are model-generated (Stage~1 answers).
    API models show near-zero discrimination, consistent with
  strong self-correction from full evidence in Stage~2.}
  \label{tab:app-history}
\end{table}

\begin{table}[ht]
  \centering
  \small
  \tablestartstriped{3}
  \begin{tabular}{@{}lrrrrrrrr@{}}
    \toprule
    \textbf{Model}
    & \textbf{MAE\textsubscript{c}}$\,\downarrow$
    & \textbf{Acc\textsubscript{10} (\%)}$\,\uparrow$
    & \textbf{UAI\textsubscript{irr}}
    & \textbf{UAI\textsubscript{pls}}
    & \textbf{TAR\textsubscript{irr}}
    & \textbf{TAR\textsubscript{pls}}
    & \textbf{Disc$_{\Delta}$}
    & \textbf{Parse} \\
    \midrule
    \qwenonefive  & 10.27 & 59.4\% & 0.01  & 0.00 & 0.23 & 0.25 & $-$0.01 & 1.00 \\
    \qwenthree    & 7.16  & 77.5\% & $-$0.04 & $-$0.04 & 0.15 & 0.21 & 0.00 & 1.00 \\
    \qwenseven    & 5.50  & 87.5\% & $-$0.03 & $-$0.01 & 0.17 & 0.23 & 0.01 & 1.00 \\
    \llamaone     & 16.26 & 54.5\% & 0.30  & 0.22 & 0.42 & 0.49 & $-$0.09 & 0.92 \\
    \llamathree   & 6.98  & 80.8\% & 0.07  & 0.14 & 0.36 & 0.44 & 0.06 & 1.00 \\
    \llamaeight   & 7.31  & 76.7\% & 0.03  & 0.15 & 0.40 & 0.51 & 0.12 & 0.97 \\
    \gemmaone     & 13.38 & 46.4\% & $-$0.00 & $-$0.03 & 0.16 & 0.24 & $-$0.02 & 1.00 \\
    \gemmafour    & 13.53 & 38.6\% & 0.00  & $-$0.06 & 0.31 & 0.33 & $-$0.07 & 1.00 \\
    \olmothirteen & 10.56 & 56.9\% & 0.03  & 0.02 & 0.18 & 0.29 & $-$0.01 & 1.00 \\
    \olmonthirtytwo & 8.15 & 69.7\% & $-$0.02 & $-$0.01 & 0.14 & 0.19 & 0.02 & 1.00 \\
    \midrule
    \gptfivefourmini & 0.47  & 99.4\% & $-$0.00 & $-$0.01 & 0.12 & 0.13 & $-$0.01 & 1.00 \\
    \claudehaiku     & 1.07  & 96.9\% & $-$0.00 & $-$0.02 & 0.11 & 0.23 & $-$0.01 & 1.00 \\
    \geminiflash     & 0.31  & 99.4\% & $-$0.00 & 0.00  & 0.06 & 0.07 & 0.00 & 1.00 \\
    \grokthreemini   & 0.39  & 99.2\% & $-$0.00 & 0.00  & 0.04 & 0.06 & 0.01 & 1.00 \\
    \bottomrule
  \end{tabular}
  \tablestopstriped
  \caption{\iclsuite~suite full results (360~items).
    Anchors appear only in demo-header metadata;
  demo evidence and answers are identical across conditions.}
  \label{tab:app-icl}
\end{table}

\begin{table}[ht]
  \centering
  \small
  \tablestartstriped{3}
  \begin{tabular}{@{}lrrrrrrrr@{}}
    \toprule
    \textbf{Model}
    & \textbf{MAE\textsubscript{c}}$\,\downarrow$
    & \textbf{Acc\textsubscript{10} (\%)}$\,\uparrow$
    & \textbf{UAI\textsubscript{irr}}
    & \textbf{UAI\textsubscript{pls}}
    & \textbf{TAR\textsubscript{irr}}
    & \textbf{TAR\textsubscript{pls}}
    & \textbf{Disc$_{\Delta}$}
    & \textbf{Parse} \\
    \midrule
    \qwenonefive  & 22.15 & 21.9\% & 0.02  & 0.23 & 0.18 & 0.49 & 0.21 & 1.00 \\
    \qwenthree    & 8.12  & 74.1\% & $-$0.02 & 0.08 & 0.15 & 0.39 & 0.10 & 1.00 \\
    \qwenseven    & 9.83  & 63.9\% & $-$0.00 & 0.20 & 0.11 & 0.48 & 0.20 & 1.00 \\
    \llamaone     & 13.73 & 60.5\% & 0.10  & 0.09 & 0.29 & 0.35 & $-$0.01 & 0.96 \\
    \llamathree   & 10.60 & 71.5\% & 0.14  & 0.22 & 0.29 & 0.36 & 0.08 & 0.99 \\
    \llamaeight   & 4.66  & 85.1\% & $-$0.02 & 0.12 & 0.22 & 0.36 & 0.14 & 0.99 \\
    \gemmaone     & 14.98 & 37.2\% & $-$0.00 & 0.04 & 0.12 & 0.23 & 0.04 & 1.00 \\
    \gemmafour    & 12.64 & 47.5\% & 0.05  & 0.01 & 0.23 & 0.32 & $-$0.05 & 1.00 \\
    \olmothirteen & 4.21  & 87.2\% & $-$0.00 & 0.05 & 0.29 & 0.35 & 0.05 & 1.00 \\
    \olmonthirtytwo & 11.00 & 58.1\% & 0.01  & 0.45 & 0.07 & 0.57 & 0.44 & 1.00 \\
    \midrule
    \gptfivefourmini & 0.57  & 98.6\% & 0.01  & 0.09 & 0.11 & 0.29 & 0.08 & 1.00 \\
    \claudehaiku     & 1.10  & 98.3\% & 0.00  & 0.06 & 0.18 & 0.31 & 0.05 & 1.00 \\
    \geminiflash     & 5.55  & 84.2\% & 0.08  & 0.17 & 0.26 & 0.36 & 0.09 & 0.99 \\
    \grokthreemini   & 0.26  & 99.2\% & $-$0.01 & 0.02 & 0.03 & 0.08 & 0.03 & 1.00 \\
    \bottomrule
  \end{tabular}
  \tablestopstriped
  \caption{\ragsuite~suite full results (360~items).}
  \label{tab:app-rag}
\end{table}

\begin{table}[ht]
  \centering
  \small
  \tablestartstriped{3}
  \begin{tabular}{@{}lrrrrrrrr@{}}
    \toprule
    \textbf{Model}
    & \textbf{MAE\textsubscript{c}}$\,\downarrow$
    & \textbf{Acc\textsubscript{10} (\%)}$\,\uparrow$
    & \textbf{UAI\textsubscript{irr}}
    & \textbf{UAI\textsubscript{pls}}
    & \textbf{TAR\textsubscript{irr}}
    & \textbf{TAR\textsubscript{pls}}
    & \textbf{Disc$_{\Delta}$}
    & \textbf{Parse} \\
    \midrule
    \qwenonefive  & 26.17 & 45.9\% & 0.26  & 0.82 & 0.28 & 0.72 & 0.57 & 0.97 \\
    \qwenthree    & 4.62  & 89.2\% & 0.08  & 0.18 & 0.24 & 0.37 & 0.10 & 0.98 \\
    \qwenseven    & 0.67  & 98.9\% & 0.00  & 0.17 & 0.08 & 0.51 & 0.17 & 1.00 \\
    \llamaone     & ---   & ---  & ---   & ---  & ---  & ---  & ---  & 0.01 \\
    \llamathree   & 2.33  & 94.5\% & 0.03  & 0.21 & 0.17 & 0.34 & 0.19 & 0.80 \\
    \llamaeight   & 12.15 & 73.3\% & 0.31  & 0.48 & 0.28 & 0.65 & 0.17 & 0.96 \\
    \gemmaone     & 20.51 & 25.8\% & $-$0.05 & $-$0.03 & 0.23 & 0.31 & 0.02 & 1.00 \\
    \gemmafour    & 14.21 & 62.8\% & 0.18  & 0.21 & 0.48 & 0.55 & 0.04 & 0.97 \\
    \olmothirteen & 9.28  & 70.9\% & 0.09  & 0.07 & 0.37 & 0.47 & $-$0.02 & 1.00 \\
    \olmonthirtytwo & 7.98 & 72.5\% & 0.01  & 0.02 & 0.11 & 0.15 & 0.01 & 1.00 \\
    \midrule
    \gptfivefourmini & 0.52  & 99.4\% & 0.00  & 0.06 & 0.13 & 0.43 & 0.06 & 1.00 \\
    \claudehaiku     & 1.27  & 98.3\% & $-$0.01 & 0.01 & 0.15 & 0.27 & 0.02 & 1.00 \\
    \geminiflash     & 1.12  & 98.3\% & 0.03  & 0.06 & 0.12 & 0.19 & 0.03 & 0.98 \\
    \grokthreemini   & 0.45  & 98.9\% & $-$0.01 & 0.02 & 0.01 & 0.11 & 0.02 & 1.00 \\
    \bottomrule
  \end{tabular}
  \tablestopstriped
  \caption{\toolsuite~suite full results (360~items).
    Tool outputs are structured (\texttt{role:~"tool"}) for
    Qwen and Llama, and plaintext for Gemma, OLMo, and API
    models.
  Llama-3.2-1B metrics are undefined (parse rate 0.9\%).}
  \label{tab:app-tool}
\end{table}

\subsection{Cross-suite patterns}
\label{app:suite-highlights}
This section summarizes cross-suite patterns that help interpret
Tables~\ref{tab:app-external}--\ref{tab:app-tool} without repeating
all cell-level values.
Table~\ref{tab:uai_summary} reports suite-level means for
UAI by relevance.

\begin{table}[ht]
  \centering
  \small
  \begin{tabular}{@{}lrrrl@{}}
    \toprule
    \textbf{Suite}
    & \textbf{UAI\textsubscript{irr}}
    & \textbf{UAI\textsubscript{pls}}
    & \textbf{Disc$_{\Delta}$}
    & \textbf{$\frac{+}{n}$} \\
    \midrule
    \externalsuite & 0.05  & 0.22 & 0.17 & 14/14 \\
    \historysuite  & 0.05  & 0.34 & 0.28 & 10/14 \\
    \iclsuite (metadata)    & 0.02  & 0.03 &  0.00 & 7/14 \\
    \iclsuite (ICL-dist)    & 0.09  & 0.16 &  0.07 & 8/10 \\
    \ragsuite      & 0.03  & 0.13 & 0.10 & 12/14 \\
    \toolsuite     & 0.07  & 0.18 & 0.11 & 12/13 \\
    \midrule
    \textit{All (excl.\ ICL-dist)} & 0.04  & 0.18 & 0.13 & 55/69 \\
    \bottomrule
  \end{tabular}
  \caption{Mean anchoring influence by relevance and suite.
    $+/n$: cells with Disc$_{\Delta} > 0$.
    ICL-dist uses distribution-matching demonstrations
  (open-weight models only).}
  \label{tab:uai_summary}
\end{table}

Anchoring concentrates in \externalsuite, \historysuite,
and \toolsuite; \iclsuite (metadata) is near zero; \ragsuite
is moderate.
History and Tool magnitudes are qualified by format confounds
(Appendices~\ref{app:history-matched}, \ref{app:tool-plaintext}).
API models show smaller but non-zero discrimination, mainly on
\externalsuite and \ragsuite.

\subsection{Statistical robustness checks}
\label{app:stats}
\paragraph{Reporting pipeline.}
Per-item UAI values are computed for each
(model, suite, relevance, direction) cell (excluding
$\varepsilon$-threshold items), then pooled across direction to
give one UAI$_{\text{irr}}$ and one UAI$_{\text{pls}}$ mean per
model--suite cell; Disc$_{\Delta}$ is their difference.
Suite-level means aggregate over models with equal weight, and
bootstrap CIs and Wilcoxon tests treat each model--suite cell as
one observation.

\paragraph{Item-level uncertainty.}
For each model--suite run, we report 95\% nonparametric bootstrap CIs
($B{=}2000$, seed~42) over items for control accuracy,
UAI components, Disc$_{\Delta}$, and parse rate.
These intervals complement the model-level summaries in
Table~\ref{tab:stats_inference}.

\begin{table}[ht]
  \centering
  \small
\begin{tabular}{@{}lrrr@{}}
    \toprule
    \textbf{Contrast} & \textbf{Est.} &
    \multicolumn{1}{c}{\textbf{95\% CI}} &
    \textbf{$p_{\mathrm{BH}}$} \\
    \midrule
    \multicolumn{4}{@{}l}{\emph{Plausible $>$ irrelevant} (mean $\Delta$UAI)} \\
    \quad \externalsuite & 0.17 & [0.13, 0.22] & $<$0.01 \\
    \quad \historysuite  & 0.28 & [0.12, 0.47] & 0.02 \\
    \quad \iclsuite      & 0.00 & [$-$0.02, 0.03] & 0.81 \\
    \quad \ragsuite      & 0.10 & [0.05, 0.17] & $<$0.01 \\
    \quad \toolsuite     & 0.11 & [0.04, 0.20] & $<$0.01 \\
    \midrule
    Suite-mean Disc$_{\Delta}$ range (open-weight panel)    & 0.40 & [0.20, 0.60] & --- \\
    Pearson $r$(Acc, Disc$_{\Delta}$)    & $-$0.24 & [$-$0.43, $-$0.00] & --- \\
    \bottomrule
  \end{tabular}

  \caption{Statistical summary.
    Top: Wilcoxon signed-rank tests for
    UAI$_{\text{pls}} > \text{UAI}_{\text{irr}}$
    ($n{=}14$ models per suite, $n{=}13$ on Tool;
    $p$-values BH-corrected).
    Bottom: cross-suite discrimination range and
    Acc--Disc correlation.
  All CIs are bootstrap 95\%.}
  \label{tab:stats_inference}
\end{table}

\paragraph{Paired cross-suite comparisons.}
History has higher Disc$_{\Delta}$ than ICL on average across models
(mean difference $0.24$, bootstrap 95\% CI $[0.11, 0.41]$;
Wilcoxon $p_{\mathrm{BH}} \approx 0.01$).
Tool Disc$_{\Delta}$ is not significantly different from the mean of
the other suites on models with defined Tool discrimination
($p_{\mathrm{BH}} \approx 0.56$).
Tool parse rate is modestly lower than the mean parse rate over
External, History, ICL, and RAG (mean gap about 12 percentage points,
bootstrap 95\% CI $[-0.31, -0.009]$; $p_{\mathrm{BH}} \approx 0.05$),
consistent with higher format sensitivity.

\paragraph{Interpretation limits.}
The model panel is not a random sample of all LLMs, so inferential
statistics should be read as internal consistency checks rather than
population claims.
Cells are also structured by model and suite, so correlation intervals
are descriptive.

\begin{figure}[ht]
  \centering
  \includegraphics[width=0.85\columnwidth]{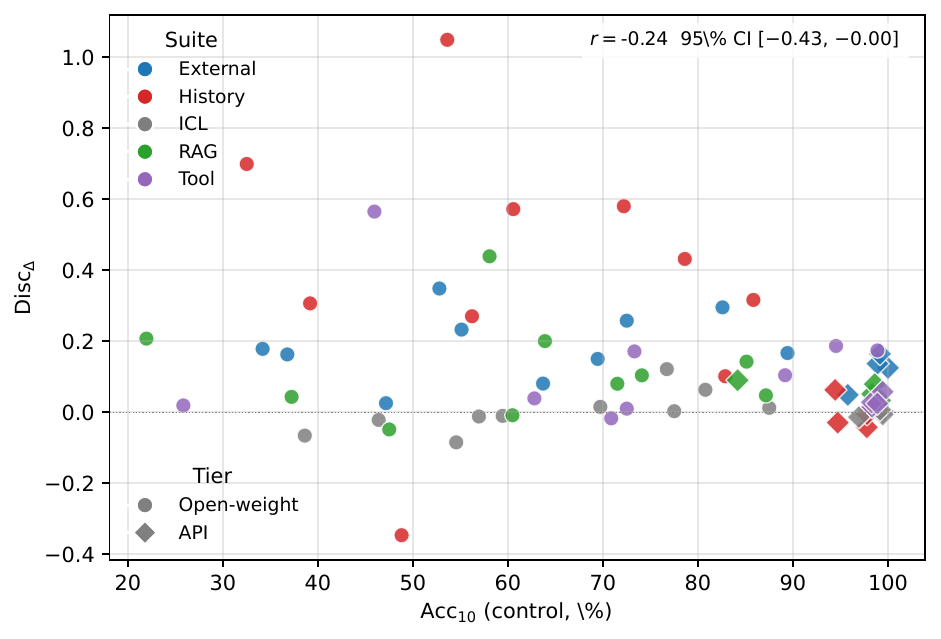}
  \caption{Task accuracy (Acc$_{10}$) vs.\ discrimination
    (Disc$_{\Delta}$) for all 69 computable model--suite cells.
    Circles: open-weight; diamonds: API; color: suite.
    The weak Pearson $r$ ($-$0.24, 95\% CI
    [$-$0.43, $-$0.00]) shows that accuracy and anchoring
  robustness are only weakly associated.}
  \label{fig:acc-vs-disc}
\end{figure}

\subsection{UAI exclusion-threshold sensitivity}
\label{app:epsilon}
The UAI metric (\S\ref{sec:metrics}) excludes items where
$|a - y_{\text{ctrl}}| < \varepsilon$ to avoid division by small
denominators.
We report exclusion rates and Disc$_{\Delta}$
stability across $\varepsilon \in \{1, 3, 5\}$.

\paragraph{Exclusion rates.}
At the default $\varepsilon{=}3$, mean exclusion rates range from
3.8\% (ICL) to 11\% (History).
History's higher rate is expected: the self-generated Stage~1
anchor can closely approximate the gold answer for high-accuracy
models.
At $\varepsilon{=}5$, rates increase further but remain
moderate.

\paragraph{Sign stability.}
Across 69 model--suite cells with computable Disc$_{\Delta}$,
most ($\approx$80\%) preserve the same sign at all three $\varepsilon$ values.
The 13 cells with sign flips are concentrated in ICL
(7 cells, where effects are near zero) and small-magnitude
RAG/Tool cells.
No External or History cell with $|\text{Disc}_{\Delta}| > 0.05$
shows a sign flip, confirming that the main findings are robust
to the exclusion threshold.

\subsection{UAI distribution and extreme values}
\label{app:uai-distribution}
UAI can exceed~1 (overshoot past the anchor) or be negative
(reverse shift).
Table~\ref{tab:uai_distribution} summarizes the distribution
across all per-item UAI values in the benchmark
(44{,}261 irrelevant, 43{,}530 plausible).

\begin{table}[ht]
  \centering
  \small
\begin{tabular}{@{}lrr@{}}
    \toprule
    & \textbf{UAI\textsubscript{irr}} & \textbf{UAI\textsubscript{pls}} \\
    \midrule
    In $[0,1]$       & 77.1\% & 73.4\% \\
    Overshoot ($>1$)  & 4.2\%  & 7.0\%  \\
    Reverse ($<0$)    & 18.7\% & 19.7\% \\
    \midrule
    Mean              & 0.04   & 0.17   \\
    Median            & 0.00   & 0.00   \\
    Capped $[0,1]$ mean & 0.10 & 0.22   \\
    \midrule
    5th pctl          & $-$0.67 & $-$0.78 \\
    25th pctl         & 0.00    & 0.00   \\
    75th pctl         & 0.00    & 0.33   \\
    95th pctl         & 0.97    & 1.32   \\
    \bottomrule
  \end{tabular}

  \caption{Distribution of per-item UAI values across all
    models and suites.
    The majority of items fall in $[0,1]$, but roughly
    4--7\% overshoot and $\sim$19\% show reverse shifts.
  Overshoot concentrates in History (23.8\% for plausible).}
  \label{tab:uai_distribution}
\end{table}

\paragraph{Robustness to aggregation method.}
The median UAI is zero for both relevance levels, reflecting
that many items show no shift.
However, the sign of
Disc$_{\Delta} = \text{UAI}_{\text{pls}} - \text{UAI}_{\text{irr}}$
is preserved for all five suites under both mean and median
aggregation, and under capped $[0,1]$ means.
Capping reduces the History Disc$_{\Delta}$ from 0.40 to 0.23
(overshoot accounts for part of the magnitude) but does not
change the direction or relative ranking of suites.

\subsection{Boundary clipping sensitivity}
\label{app:boundary}
At $\delta{=}40$, some anchors are clipped to the scale
boundaries ($a{=}0$ or $a{=}100$) when $\theta < 40$ or
$\theta > 60$.
To check whether this clipping inflates the apparent
attenuation at large offsets, we compare dose-response trends
with and without boundary items (External + RAG, plausible
UAI):
\begin{center}
  \small
  \begin{tabular}{@{}lccc@{}}
    \toprule
    & $\delta{=}15$ & $\delta{=}25$ & $\delta{=}40$ \\
    \midrule
    All items           & 0.23 & 0.18 & 0.11 \\
    Excl.\ boundary     & 0.23 & 0.18 & 0.14 \\
    \bottomrule
  \end{tabular}
\end{center}
Excluding 1{,}731 boundary items (26\% of $\delta{=}40$ items)
raises the mean UAI$_{\text{pls}}$ at $\delta{=}40$ from 0.11
to 0.14.
The monotonic attenuation trend is preserved, but roughly
half of the drop from $\delta{=}25$ to $\delta{=}40$ is
attributable to boundary clipping.
At $\delta \in \{15, 25\}$, no items hit the boundaries
(all $\theta \in [30,70]$).

\subsection{Anchored-condition prediction quality}
\label{app:anchored-mae}
We measure whether anchored conditions degrade prediction
quality by comparing MAE in anchored vs.\ control conditions.
Table~\ref{tab:anchored_mae} reports the mean
$\Delta\text{MAE} = \text{MAE}_{\text{anchored}} - \text{MAE}_{\text{ctrl}}$
across models.

\begin{table}[ht]
  \centering
  \small
\begin{tabular}{@{}lrr@{}}
    \toprule
    \textbf{Suite}
    & \textbf{$\Delta$MAE\textsubscript{irr}}
    & \textbf{$\Delta$MAE\textsubscript{pls}} \\
    \midrule
    \externalsuite & $+$0.08 & $+$3.61 \\
    \historysuite  & $-$1.60 & $+$0.12 \\
    \iclsuite      & $+$0.14 & $-$0.04 \\
    \ragsuite      & $-$0.04 & $+$0.59 \\
    \toolsuite     & $-$0.67 & $+$3.39 \\
    \bottomrule
  \end{tabular}

  \caption{Mean change in MAE from control to anchored
    conditions (averaged across all models per suite).
    Positive values indicate accuracy degradation.
    Plausible anchors consistently increase MAE;
    irrelevant anchors are generally small,
  though Tool shows a modest negative effect ($-$0.67).}
  \label{tab:anchored_mae}
\end{table}

\noindent Plausible anchors increase MAE by 0.6--3.6 on External,
RAG, and Tool, confirming that anchoring produces \emph{substantive}
accuracy degradation, not just directional shifts.
History shows a near-zero plausible effect (+0.12) but a notable
negative irrelevant shift ($-$1.61), likely an artifact of
single-stage control noise (\S\ref{app:history-matched}).
The negative $\Delta\text{MAE}_{\text{irr}}$ for Tool ($-$0.65)
may reflect that irrelevant-condition prompts contain structured
tool output that contextualizes the task, marginally improving
predictions for some models despite the irrelevant anchor.
ICL shows near-zero $\Delta\text{MAE}$ for both relevance levels,
consistent with the weak manipulation.

\subsection{History suite matched-control analysis}
\label{app:history-matched}
The default History control is single-stage while anchored
conditions use a two-stage protocol (\S\ref{sec:suites}).
To assess whether format differences confound anchoring
estimates, we compare the standard Disc$_{\Delta}$ against a
\emph{matched-control} Disc$_{\Delta}$ computed using a
two-stage control baseline (where Stage~1 presents a different
  case, introducing the two-stage format without an
item-specific anchor).

\begin{table}[ht]
  \centering
  \small
  \tablestartstriped{3}
  \begin{tabular}{@{}lrrrr@{}}
    \toprule
    \textbf{Model}
    & \textbf{MAE\textsubscript{std}}
    & \textbf{MAE\textsubscript{ts}}
    & \textbf{Disc\textsubscript{std}}
    & \textbf{Disc\textsubscript{match}} \\
    \midrule
    \qwenonefive  & 5.49  & 24.87 & 0.89 & 0.27 \\
    \qwenthree    & 3.65  & 17.67 & 0.18 & 0.51 \\
    \qwenseven    & 3.77  & 10.33 & 0.18 & 0.97 \\
    \llamaone  & 15.71 & 20.87 & 0.01 & 0.39 \\
    \llamathree  & 9.46  & 15.42 & 0.22 & 0.20 \\
    \llamaeight  & 7.02  & 8.05  & 0.08 & 0.33 \\
    \gemmaone & 14.71 & 27.45 & 0.21 & 0.09 \\
    \gemmafour & 12.03 & 17.32 & 0.24 & 0.57 \\
    \olmothirteen  & 5.40  & 9.86  & 0.29 & 0.82 \\
    \olmonthirtytwo  & 4.13  & 8.51  & 0.12 & 0.69 \\
    \bottomrule
  \end{tabular}
  \tablestopstriped
  \caption{History suite: standard vs.\ matched-format
    (two-stage) control analysis.
    MAE\textsubscript{std}: MAE on single-stage control;
    MAE\textsubscript{ts}: MAE on two-stage control;
    Disc\textsubscript{std}: Disc$_{\Delta}$ using single-stage control;
    Disc\textsubscript{match}: Disc$_{\Delta}$ using two-stage control.
    Both columns are computed from the dedicated
    matched-control evaluation run and should be compared
    against each other; Disc\textsubscript{std} values may
    differ from Table~\ref{tab:main_results} because the
    matched-control run used a separate item subset and
  inference pass.}
  \label{tab:history_matched}%
  \label{tab:history_dual}
\end{table}

\paragraph{Key observations.}
The two-stage control format substantially increases MAE
(for example, 5.49 $\to$ 24.87 for Qwen-1.5B;
mean across models: 8.1 $\to$ 16.0), confirming the
format itself adds difficulty.
Disc$_{\Delta}$ drops substantially for 2/10 models:
Qwen-1.5B drops from 0.89 to 0.27 (the standard estimate was
severely inflated by the format confound),
while Gemma-1B drops from 0.21 to 0.09.
Most other models show \emph{increased} Disc$_{\Delta}$
under matched control, likely because the two-stage control
is noisier (higher MAE), making the shift toward anchors
relatively larger.

\paragraph{Interpretation.}
The sign of Disc$_{\Delta}$ is preserved for 10/10 models,
supporting genuine anchoring above format effects.
However, the magnitudes vary substantially between the two
estimates.
We recommend interpreting History Disc$_{\Delta}$ as
qualitative evidence for self-generated anchoring rather than
as precise quantitative estimates, and note that the standard
(single-stage control) likely overestimates the effect for
small models with large format sensitivity.

\subsection{ICL distribution-matching comparison}
\label{app:icl-dist}
The standard ICL suite uses metadata anchors (case IDs, batch
numbers) that are transparently irrelevant to the estimation
task (\S\ref{sec:suites}).
To test whether stronger numeric priming produces measurable
effects, we create a \emph{distribution-matching} ICL variant
where few-shot demonstration answers are numerically relevant
to the target range.

\begin{table}[ht]
  \centering
  \small
  \tablestartstriped{3}
  \begin{tabular}{@{}lrrrr@{}}
    \toprule
    \textbf{Model}
    & \textbf{UAI\textsubscript{pls,std}}
    & \textbf{UAI\textsubscript{pls,dist}}
    & \textbf{Disc\textsubscript{std}}
    & \textbf{Disc\textsubscript{dist}} \\
    \midrule
    \qwenonefive  &  0.00 &  0.43 & $-$0.01 &  0.09 \\
    \qwenthree    & $-$0.04 &  0.15 &  0.00 &  0.07 \\
    \qwenseven    & $-$0.01 &  0.06 &  0.01 &  0.07 \\
    \llamaone  &  0.22 &  0.22 & $-$0.09 &  0.04 \\
    \llamathree  &  0.14 &  0.11 &  0.06 &  0.10 \\
    \llamaeight  &  0.15 &  0.26 &  0.12 & $-$0.03 \\
    \gemmaone & $-$0.03 &  0.00 & $-$0.02 & $-$0.03 \\
    \gemmafour & $-$0.06 &  0.12 & $-$0.07 &  0.19 \\
    \olmothirteen  &  0.02 &  0.12 & $-$0.01 &  0.12 \\
    \olmonthirtytwo  & $-$0.01 &  0.13 &  0.02 &  0.09 \\
    \bottomrule
  \end{tabular}
  \tablestopstriped
  \caption{Standard ICL (metadata anchors) vs.\ ICL-dist
  (numerically relevant demonstrations).}
  \label{tab:icl_dist}
\end{table}

\paragraph{Results.}
Eight of ten models show increased UAI$_{\text{pls}}$ under
the distribution-matching variant
(mean: 0.04 $\to$ 0.16, paired $t$-test $p < 0.05$).
The largest increase is for Qwen-1.5B
(0.00 $\to$ 0.43), consistent with smaller models being more
susceptible to numeric priming in demonstrations.
The distribution-matching variant also produces
Disc$_{\Delta} > 0$ for 8/10 models
(vs.\ 7/10 for standard ICL), confirming that relevance
discrimination emerges when the anchoring pathway is
sufficiently strong.

\paragraph{Interpretation.}
The near-zero standard ICL result reflects the \emph{weakness
of the manipulation} (incidental metadata numbers) rather than
model immunity to few-shot anchoring.
When demonstrations contain numerically relevant information,
models show substantial anchoring comparable to RAG and Tool
suites.
This supports the design choice to include ICL as a
minimal-exposure baseline while demonstrating that stronger
ICL manipulations produce expected effects.

\subsection{Tool suite format comparison}
\label{app:tool-plaintext}
The Tool suite uses structured \texttt{role:~"tool"} messages for
Qwen and Llama models but plaintext context for Gemma, OLMo, and
API models (\S\ref{sec:suites}).
To isolate format effects, we re-evaluate all five
structured-format models with tool outputs rendered as
plaintext (identical to the Gemma/OLMo/API format).

\begin{table}[ht]
  \centering
  \small
  \tablestartstriped{3}
  \begin{tabular}{@{}lrrrr@{}}
    \toprule
    \textbf{Model}
    & \textbf{Disc\textsubscript{struct}}
    & \textbf{Disc\textsubscript{plain}}
    & \textbf{MAE\textsubscript{struct}}
    & \textbf{MAE\textsubscript{plain}} \\
    \midrule
    \qwenonefive  & 0.58  & 0.07  & 11.34 & 15.01 \\
    \qwenthree    & 0.08  & 0.11  & 2.19  & 15.74 \\
    \qwenseven    & 0.17  & 0.02  & 0.71  & 1.32 \\
    \llamathree  & 0.23  & 0.10  & 2.28  & 9.49 \\
    \llamaeight  & 0.16  & 0.14  & 5.47  & 6.77 \\
    \bottomrule
  \end{tabular}
  \tablestopstriped
  \caption{Tool suite: structured (\texttt{role:~"tool"}) vs.\
    plaintext format for all five models that used structured
    format in the main benchmark.
  Disc: Disc$_{\Delta}$; MAE: control-condition MAE.}
  \label{tab:tool_plaintext}
\end{table}

\paragraph{Results.}
Format effects are substantial and model-dependent.
Three of five models show reduced discrimination in plaintext:
Qwen-1.5B drops from 0.58 to 0.07 (the largest change in the
benchmark), Qwen-7B from 0.17 to 0.02, and Llama-3B from 0.23
to 0.10.
Two models are stable: Qwen-3B remains near 0.08--0.11 and
Llama-8B retains 0.14--0.16.
MAE increases in plaintext mode for all models, confirming
that the structured format provides better task context.

\paragraph{Implications.}
The structured tool format amplifies both task competence
(lower MAE) and anchor sensitivity (higher Disc) for some
models, particularly smaller Qwen models.
Cross-model Tool-suite comparisons in Table~\ref{tab:main_results}
mix formats (structured for Qwen/Llama; plaintext for
Gemma/OLMo/APIs) and should be interpreted accordingly.
The External and RAG suites, which use identical input formats
across all models, provide cleaner cross-model comparisons.

\subsection{Gold-referenced error decomposition}
\label{app:gold-shift}

Table~\ref{tab:gold_shift_main} reports the aggregate
decomposition of anchored responses into error-increasing,
error-reducing, and neutral shifts relative to the evidence-only
gold answer; this section adds the per-model breakdown and the
anchor-proximity stratification.

\begin{table}[ht]
  \centering
  \small
  \begin{tabular}{@{}lrrrr@{}}
    \toprule
    \textbf{Suite}
    & \textbf{Err$\uparrow$\,(\%)}
    & \textbf{Err$\downarrow$\,(\%)}
    & \textbf{Irr\textsubscript{err$\uparrow$}}
    & \textbf{Pls\textsubscript{err$\uparrow$}} \\
    \midrule
    \externalsuite & 33.9 & 20.0 & 19.9 & 48.0 \\
    \historysuite  & 39.2 & 44.0 & 29.5 & 48.7 \\
    \iclsuite      & 22.9 & 20.0 & 20.3 & 25.4 \\
    \ragsuite      & 22.2 & 20.2 & 16.4 & 27.9 \\
    \toolsuite     & 28.6 & 17.5 & 17.6 & 39.9 \\
    \midrule
    \textit{All}   & 28.7 & 23.2 & 20.2 & 37.3 \\
    \bottomrule
  \end{tabular}
  \caption{Error decomposition relative to the evidence-only
    target (aggregated over fourteen models).
    Err$\uparrow$/Err$\downarrow$: share of anchored items where the
    response shifts away from / toward the gold answer.
    Irr\textsubscript{err$\uparrow$}/Pls\textsubscript{err$\uparrow$}:
    error-increasing rate by anchor relevance.
    Overall, error-increasing shifts (29\%) outnumber
    error-reducing ones (23\%); the gap is driven by plausible
    anchors (37\% vs.\ 20\% for irrelevant), with the largest
  differences on External and Tool ($\ge$22~pp).}
  \label{tab:gold_shift_main}
\end{table}

\begin{figure}[ht]
  \centering
  \includegraphics[width=\columnwidth]{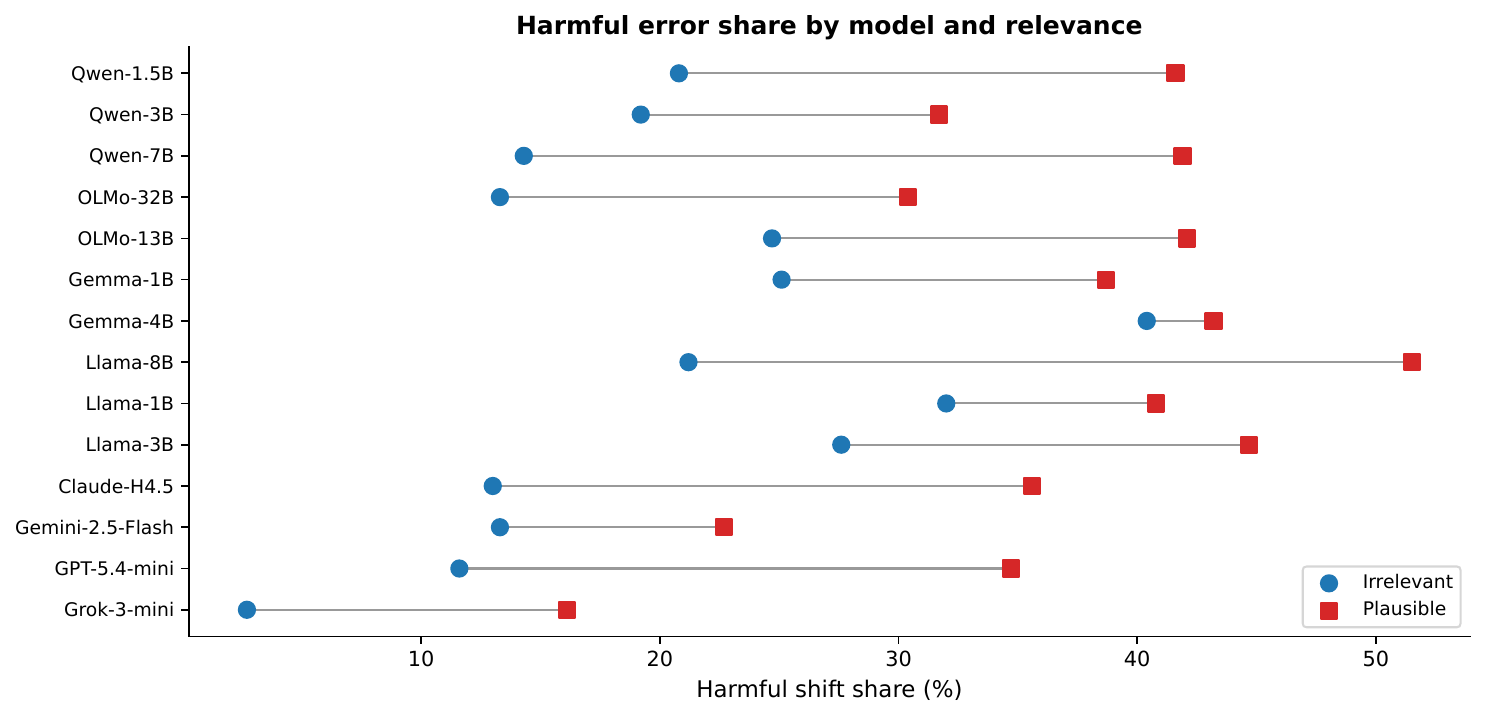}
  \caption{Harmful shift share by model and relevance condition.
    Each point shows the percentage of anchored items on which
    the model's error increased relative to the unanchored
    control.
    Plausible anchors (right) consistently produce a larger
    harmful share than irrelevant anchors (left), with the
    widest gaps for External and Tool suites.}
  \label{fig:gold-shift-per-model}
\end{figure}

\paragraph{Suite-level patterns.}
External shows the sharpest relevance contrast:
plausible anchors produce harmful shifts 48.0\% of the time
versus 19.9\% for irrelevant anchors ($\Delta = 28$~pp).
Tool follows with a large gap (39.9\% vs.\ 17.6\%, $\Delta = 22$~pp).
RAG shows a moderate gap (27.9\% vs.\ 16.4\%).
ICL shows modest differentiation between relevance levels
(25.4\% vs.\ 20.3\%), consistent with the weak
discrimination in the standard ICL suite.
History shows a 19~pp gap (48.7\% vs.\ 29.5\%).

\paragraph{Proximity analysis.}
When the anchor is farther from the gold answer than the
control response, 31.0\% of items show harmful shifts.
When the anchor is closer to gold, only 14.9\% are harmful.
This directional asymmetry is expected: anchors that pull
responses away from gold are more likely to increase error,
while anchors that lie between the control and gold may
coincidentally improve predictions.
Among the ``closer'' items, helpful shifts are common
(59.3\%), confirming that the helpful category largely
reflects cases where the anchor direction happens to
align with gold.

\paragraph{Interpretation.}
The decomposition shows that plausible anchors increase
prediction error more often than they reduce it
($37.3\%$ vs.\ $24.1\%$; the remaining $38.6\%$ of items are
unchanged), and substantially more often than irrelevant
anchors.
The 17~pp gap between plausible and irrelevant harmful
rates ($37.3\%$ vs.\ $20.2\%$) mirrors the discrimination
observed in the main UAI analysis and confirms that the
anchor-induced shifts are not symmetric noise but
directionally biased toward the anchor value.

\paragraph{Breakdown by difficulty and offset.}
Hard items produce more non-neutral shifts overall:
harmful~31.6\% vs.\ 25.8\% for easy items, and
helpful~25.4\% vs.\ 21.1\%.
This is consistent with noisier evidence leaving more room
for anchor-driven shifts in either direction.
Across anchor offsets, plausible harmful rates are
similar at $\delta{=}15$ (35.5\%) and $\delta{=}25$ (36.1\%)
but decline at $\delta{=}40$ (33.9\%), while irrelevant
harmful rates remain stable (${\approx}$18.4\%).
The moderate decline at $\delta{=}40$ is consistent
with the dose-response attenuation, partly amplified
by boundary clipping (Appendix~\ref{app:boundary}).

\subsection{Difficulty-conditioned analysis}
\label{app:difficulty}
Table~\ref{tab:difficulty} reports anchoring metrics
separately for \emph{easy} items ($\sigma{=}8$, all five
ratings visible) and \emph{hard} items ($\sigma{=}15$, two
hidden, one conflicting).

\begin{table}[ht]
  \centering
  \small
  \setlength{\tabcolsep}{3.5pt}
\begin{tabular}{@{}llrrrrr@{}}
    \toprule
    \textbf{Suite} & \textbf{Diff.}
    & \textbf{Acc$_{10}$}
    & \textbf{MAE\textsubscript{c}}
    & \textbf{UAI\textsubscript{irr}}
    & \textbf{UAI\textsubscript{pls}}
    & \textbf{Disc$_{\Delta}$} \\
    \midrule
    \externalsuite & easy & 77.0\% & 7.2 & 0.05 & 0.16 & 0.11 \\
    & hard & 65.5\% & 9.6 & 0.04 & 0.28 & \textbf{0.23} \\
    \historysuite  & easy & 77.1\% & 7.5 & 0.03 & 0.36 & \textbf{0.33} \\
    & hard & 65.1\% & 10.1 & 0.08 & 0.31 & 0.23 \\
    \iclsuite      & easy & 76.7\% & 6.7 & 0.03 & 0.02 & $-$0.01 \\
    & hard & 72.4\% & 7.8 & 0.02 & 0.03 & \textbf{0.01} \\
    \ragsuite      & easy & 77.7\% & 6.8 & 0.02 & 0.07 & 0.05 \\
    & hard & 63.3\% & 10.3 & 0.03 & 0.19 & \textbf{0.16} \\
    \toolsuite     & easy & 79.0\% & 8.0 & 0.09 & 0.17 & 0.08 \\
    & hard & 79.2\% & 7.6 & 0.05 & 0.18 & \textbf{0.13} \\
    \midrule
    \textit{All}   & easy & --- & --- & 0.04 & 0.16 & 0.12 \\
    & hard & --- & --- & 0.05 & 0.20 & \textbf{0.15} \\
    \bottomrule
  \end{tabular}

  \caption{Anchoring metrics by item difficulty
    (mean across 14 models per suite, 13 on Tool).
    Bold highlights the higher Disc$_{\Delta}$ within each suite.
    Hard items show stronger discrimination on External
  and RAG; History shows the opposite pattern.}
  \label{tab:difficulty}
\end{table}

\paragraph{Findings.}
Hard items show higher overall susceptibility (mean
Disc$_{\Delta}$: 0.15 vs.\ 0.12 for easy items), driven by
External (0.23 vs.\ 0.11) and RAG (0.16 vs.\ 0.06).
This is consistent with the expectation that models are more
vulnerable to anchors when the evidence is sparser or more
conflicting.
History shows the opposite pattern (easy: 0.33 vs.\ hard:
0.23): easy items produce a stronger Stage~1 answer that
serves as a clearer self-generated anchor, increasing
susceptibility despite high task accuracy.
ICL remains near zero for both difficulty levels.

\subsection{Sampling robustness}
\label{app:sampling}

All main results use greedy decoding (temperature~0).
To assess whether conclusions depend on the decoding strategy,
we re-evaluate two models (Qwen-7B, Llama-8B) on three suites
(External, RAG, ICL-dist) under stochastic sampling
(temperature~0.7, top-$p$~0.9) with three seeds.

\begin{table}[ht]
  \centering
  \small
  \setlength{\tabcolsep}{3.5pt}
  \begin{tabular}{@{}llccc@{}}
    \toprule
    \textbf{Suite} & \textbf{Model} &
    \textbf{Disc$_{\Delta}$\textsubscript{greedy}} &
    \textbf{Disc$_{\Delta}$\textsubscript{sampled}} &
    \textbf{Sign} \\
    \midrule
    \externalsuite
    & Qwen-7B  & 0.26  & 0.29\;$\pm$\;0.02 & \cmark \\
    & Llama-8B & 0.30  & 0.24\;$\pm$\;0.07 & \cmark \\
    \ragsuite
    & Qwen-7B  & 0.19  & 0.18\;$\pm$\;0.02 & \cmark \\
    & Llama-8B & 0.16  & 0.06\;$\pm$\;0.04 & \cmark \\
    ICL-dist
    & Qwen-7B  & 0.07  & 0.08\;$\pm$\;0.02 & \cmark \\
    & Llama-8B & 0.03  & 0.00\;$\pm$\;0.02 & $\times$ \\
    \bottomrule
  \end{tabular}
  \caption{Sampling robustness.
    Sampled values are mean $\pm$ SD over 3 seeds
    (temperature 0.7, top-$p$ 0.9).
    Sign: whether greedy and sampled mean share the
  same Disc$_{\Delta}$ sign.}
  \label{tab:sampling_robustness}
\end{table}

\paragraph{Findings.}
Qwen-7B produces nearly identical metrics under sampling
(SD $\leq$ 0.02).
Llama-8B is more variable (SD up to 0.07) but preserves
the sign of Disc$_{\Delta}$ in 5 of 6 cells.
The exception---ICL-dist, where greedy yields a small
positive Disc$_{\Delta}$ (0.03) that averages to zero
under sampling---involves a near-zero-effect cell where
sign instability is expected.

Anchoring conclusions from the greedy main results are
robust: plausible $>$ irrelevant anchoring holds across
decoding strategies for all suites with meaningful effect sizes.

\subsection{Mitigation headroom probe}
\label{app:mitigation-headroom}

We test whether simple prompt-based strategies can attenuate
anchoring effects.
Three strategies are compared against the unmodified baseline
on the External and RAG suites for two open-weight models:

\begin{itemize}
  \item \textbf{Ignore:} an instruction to disregard
    extraneous numbers and base the answer only on the
    task evidence.
  \item \textbf{Self-check:} after producing an initial
    estimate, the model is asked to verify whether any
    extraneous numbers may have biased its answer, and
    correct if so.
  \item \textbf{CoT (chain-of-thought):} the model is asked
    to list the relevant evidence, compute an estimate from
    that evidence only, then state its final answer.
\end{itemize}

\begin{table}[ht]
  \centering
  \small
  \setlength{\tabcolsep}{3pt}
  \begin{tabular}{@{}llrrrr@{}}
    \toprule
    \textbf{Suite} & \textbf{Strategy}
    & \multicolumn{2}{c}{\textbf{Disc$_{\Delta}$}}
    & \multicolumn{2}{c}{\textbf{MAE\textsubscript{ctrl}}} \\
    \cmidrule(lr){3-4}\cmidrule(lr){5-6}
    & & Qwen-7B & Llama-8B & Qwen-7B & Llama-8B \\
    \midrule
    \externalsuite
    & baseline    & 0.26 & 0.30 & 7.7 & 5.5 \\
    & ignore      & 0.16 & 0.10 & 5.0 & 3.4 \\
    & self-check  & 0.45 & 0.28 & 8.8 & 7.5 \\
    & CoT         & 0.14 & 0.31 & 1.1 & 6.1 \\
    \midrule
    \ragsuite
    & baseline    & 0.20 & 0.16 & 9.8 & 4.8 \\
    & ignore      & 0.14 & 0.04 & 6.8 & 4.4 \\
    & self-check  & 0.30 & 0.11 & 12.2 & 6.0 \\
    & CoT         & 0.03 & 0.08 & 1.3 & 4.5 \\
    \bottomrule
  \end{tabular}
  \caption{Mitigation headroom probe.
    Disc$_{\Delta}$ and control MAE for three prompt-based
    strategies vs.\ unmodified baseline.
    \emph{Ignore} consistently reduces discrimination across both
    models and suites.
    \emph{CoT} strongly helps Qwen-7B but is inconsistent for
    Llama-8B.
    \emph{Self-check} increases discrimination for Qwen-7B on both
  suites but slightly reduces it for Llama-8B.}
  \label{tab:mitigation_headroom}
\end{table}

\paragraph{Results.}
The \emph{ignore} strategy consistently reduces
Disc$_{\Delta}$ across all four model--suite cells
(mean reduction: $-$0.12), with concurrent MAE improvement
on External for both models.
\emph{CoT} strongly reduces discrimination for Qwen-7B
($-$0.12 on External, $-$0.17 on RAG) and moderately for
Llama-8B on RAG ($-$0.08), but slightly increases it for
Llama-8B on External ($+$0.01).
The \emph{self-check} strategy is model-dependent:
it increases Disc$_{\Delta}$ for Qwen-7B on both suites
(External: $+$0.19; RAG: $+$0.10) while slightly reducing
it for Llama-8B (External: $-$0.01; RAG: $-$0.05).
Differences of this size should not be read directionally.
Re-running the Llama-8B$\times$External cell under an identical
configuration moves Disc$_{\Delta}$ by up to $0.06$ and
UAI$_{\text{pls}}$ by up to $0.08$, because batched GPU inference
is not bitwise reproducible even at temperature~0
(Appendix~\ref{app:cot-extended}, footnote~\ref{fn:cot-variance}).
Only the larger effects ($|\Delta| \gtrsim 0.10$) in this table
are resolved by a single run.

\paragraph{Interpretation.}
Simple prompt-based mitigations can partially attenuate
anchoring, but no single strategy eliminates it and
effectiveness is model-dependent.
The \emph{ignore} strategy is the most reliable, reducing
discrimination by 30--75\% on External and RAG.
\emph{CoT} is highly effective for Qwen-7B (which also sees
large MAE improvements, down to 1.1--1.3) but inconsistent
for Llama-8B.
\emph{Self-check} backfires for Qwen-7B, likely because
revisiting the context gives the anchor a second opportunity
to bias the response; for Llama-8B, the effect is small.
These results suggest headroom for targeted interventions but
indicate that robust debiasing will require more than
instruction-level changes.

\subsection{Additional analyses}
\label{app:additional}

This section collects supplementary analyses that probe the
robustness, interpretation, and generality of the main findings.
All use the standardized inference pipeline of
\S\ref{sec:setup} and, unless noted, the locked four-model
open-weight panel (Gemma-4B, Llama-8B, OLMo-13B, Qwen-7B), chosen
to span families and accuracy levels at manageable cost.

\subsubsection{A rational-updating reference for UAI}
\label{app:bayesian}

A bare UAI value does not say whether a shift is excessive.
We give it a reference point with a simple Gaussian-conjugate
Bayesian model: treat the visible evidence as $n$ ratings of unit
credibility and the anchor as worth $w$ ratings.
A rational updater then closes a fraction $w/(n{+}w)$ of the
control--anchor gap, so with $n{=}5$ and a reference weight
$w{=}1$ the rational UAI ceiling is $0.167$.
For an irrelevant anchor the rational weight is $0$, so the
ceiling is $0$ and any positive UAI$_{\text{irr}}$ is bias.
For a plausible anchor, we invert the relation to read off the
\emph{implied weight} $w_{\text{imp}} = \text{UAI}/(1-\text{UAI})\cdot n$,
the smallest weight that would make the observed shift rational
(Table~\ref{tab:implied_weight}).
Table~\ref{tab:excess_uai} reports the matching excess
over the ceiling.

\begin{table}[t]
\centering
\small
\setlength{\tabcolsep}{4pt}
\begin{tabular}{l r r r r r}
\toprule
Model & External & History & Rag & Tool & Icl \\
\midrule
Claude-H4.5 & 0.71 & 0.00 & 0.29 & 0.03 & 0.00 \\
GPT-5.4-mini & 0.79 & 0.23 & 0.47 & 0.33 & 0.00 \\
Gemini-2.5-Flash & 0.38 & 0.00 & 1.03 & 0.29 & 0.00 \\
Gemma-1B & 1.10 & 0.76 & 0.22 & 0.00 & 0.00 \\
Gemma-4B & 3.74$^{*}$ & 5.00$^{*}$ & 0.03 & 1.37 & 0.00 \\
Grok-3-mini & 0.97 & 0.53 & 0.11 & 0.08 & 0.02 \\
Llama-1B & 0.91 & 0.79 & 0.51 & --- & 1.38 \\
Llama-3B & 2.09$^{*}$ & 13.32$^{*}$ & 1.43 & 1.35 & 0.78 \\
Llama-8B & 2.82$^{*}$ & 1.52 & 0.69 & 4.58$^{*}$ & 0.88 \\
OLMo-13B & 0.94 & 2.30$^{*}$ & 0.25 & 0.40 & 0.08 \\
OLMo-32B & 3.27$^{*}$ & 15.17$^{*}$ & 4.04$^{*}$ & 0.10 & 0.00 \\
Qwen-1.5B & 1.25 & 319.68$^{*}$ & 1.46$^{*}$ & 23.51$^{*}$ & 0.02 \\
Qwen-3B & 0.50 & 4.88$^{*}$ & 0.43 & 1.13 & 0.00 \\
Qwen-7B & 1.86$^{*}$ & 2.99$^{*}$ & 1.22 & 1.05 & 0.00 \\
\bottomrule
\end{tabular}
\caption{Implied anchor weight $w_{\mathrm{imp}}$ on \emph{plausible} anchors, defined as the smallest weight under a Gaussian-conjugate Bayesian model that would make the observed mean UAI rational (with $n=5$ evidence ratings). Under the reference assumption that an anonymous anchor is worth one evidence item ($w=1$), the rational UAI ceiling is 0.167; $w_{\mathrm{imp}}>1$ means the model treats the anchor as more credible than a single evidence item. $w_{\mathrm{imp}}>n$ means the model treats the anchor as more credible than all visible evidence combined. $^{*}$ marks cells where the 95\% CI for UAI$_{\mathrm{pls}}$ lies entirely above the rational ceiling.}
\label{tab:implied_weight}
\end{table}

\begin{table}[t]
\centering
\small
\setlength{\tabcolsep}{4pt}
\begin{tabular}{l r r r r r}
\toprule
Model & External & History & Rag & Tool & Icl \\
\midrule
Claude-H4.5 & -0.042 & -0.226 & -0.112 & -0.161 & -0.184 \\
GPT-5.4-mini & -0.030 & -0.123 & -0.080 & -0.105 & -0.173 \\
Gemini-2.5-Flash & -0.096 & -0.171 & +0.005 & -0.111 & -0.166 \\
Gemma-1B & +0.013 & -0.034 & -0.125 & -0.195 & -0.193 \\
Gemma-4B & +0.261$^{*}$ & +0.333$^{*}$ & -0.161 & +0.048 & -0.228 \\
Grok-3-mini & -0.004 & -0.071 & -0.145 & -0.151 & -0.163 \\
Llama-1B & -0.013 & -0.031 & -0.074 & --- & +0.050 \\
Llama-3B & +0.128$^{*}$ & +0.560$^{*}$ & +0.055 & +0.046 & -0.031 \\
Llama-8B & +0.194$^{*}$ & +0.067 & -0.045 & +0.311$^{*}$ & -0.018 \\
OLMo-13B & -0.008 & +0.149$^{*}$ & -0.120 & -0.093 & -0.150 \\
OLMo-32B & +0.228$^{*}$ & +0.585$^{*}$ & +0.280$^{*}$ & -0.148 & -0.174 \\
Qwen-1.5B & +0.033 & +0.818$^{*}$ & +0.060$^{*}$ & +0.658$^{*}$ & -0.162 \\
Qwen-3B & -0.076 & +0.327$^{*}$ & -0.088 & +0.018 & -0.204 \\
Qwen-7B & +0.104$^{*}$ & +0.207$^{*}$ & +0.030 & +0.007 & -0.180 \\
\bottomrule
\end{tabular}
\caption{Excess UAI on plausible anchors above the rational Bayesian ceiling 0.167 (assuming the anchor is worth one evidence item, $w=1$, $n=5$ evidence ratings). Positive values indicate the model shifted further toward the anchor than a rational Bayesian update could justify. $^{*}$ marks cells where the 95\% CI for UAI$_{\mathrm{pls}}$ lies entirely above the ceiling.}
\label{tab:excess_uai}
\end{table}

Across the $55$ non-ICL model--suite cells, $16$ have a
UAI$_{\text{pls}}$ whose 95\% CI lies entirely above $0.167$, and
$5$ imply $w_{\text{imp}}>n{=}5$---treating one anonymous anchor
as more credible than all five ratings combined.
These are the cells we flag as practically significant in
\S\ref{sec:finding-ceiling}; the rest sit inside the band that a
rational update at $w{=}1$ would produce.

\subsubsection{Plausibility spectrum and a placebo floor}
\label{app:spectrum}

To separate anchoring on the numeric value from rational use of
source credibility, we extend the binary relevance axis to a
four-point spectrum on External: \emph{placebo} (the number is an
extraneous property such as a document's age in days),
\emph{irrelevant} (a case number), \emph{plausible}
(``a recent report suggested ${\sim}23$''), and \emph{authority}
(``a panel of senior analysts estimated ${\sim}23$'').
A purely rational model would ignore placebo and irrelevant
values equally and move only for plausible/authority framings.

\begin{table}[t]
\centering
\small
\setlength{\tabcolsep}{4pt}
\begin{tabular}{l rrrr}
\toprule
Model & UAI$_{\mathrm{placebo}}$ & UAI$_{\mathrm{irr}}$ & UAI$_{\mathrm{pls}}$ & UAI$_{\mathrm{auth}}$ \\
\midrule
GPT-5.4-mini & -0.01 & +0.01 & +0.13 & +0.40 \\
Gemma-4B & +0.38 & +0.20 & +0.43 & +0.58 \\
Llama-8B & +0.08 & +0.07 & +0.36 & +0.42 \\
OLMo-13B & +0.02 & -0.01 & +0.16 & +0.49 \\
Qwen-7B & -0.00 & +0.01 & +0.27 & +0.56 \\
\midrule
Mean & \textbf{+0.09} & \textbf{+0.05} & \textbf{+0.27} & \textbf{+0.49} \\
\bottomrule
\end{tabular}
\caption{Plausibility spectrum on External: mean UAI for each of four anchor types --- placebo (anchor refers to an extraneous numeric property such as document length), irrelevant (case-number anchor), plausible (``a recent report suggested $\sim$23''), and authority (``a panel of senior analysts estimated $\sim$23''). A monotone increase along the plausibility axis would be consistent with rational use of source credibility; any positive UAI on the placebo column is unambiguously anchoring bias.}
\label{tab:plausibility_spectrum}
\end{table}

Two observations argue against a purely rational reading.
The placebo column is not zero (mean UAI $0.09$) and sits close
to irrelevant ($0.05$)---a shift toward a number that is
literally a document length.
And the increase along the axis is not clean: for some models the
placebo shift rivals the plausible one.
This is the behavior we expect if the numeric value itself is
salient, not only its stated source.

\subsubsection{Uncertain judgment under partial evidence}
\label{app:uncertain}

The main task shows all five ratings, so the gold answer is a
fully determined mean.
To introduce genuine epistemic uncertainty, we re-render External
items showing only $k$ of the five ratings while still scoring
against the full-five mean.
Now the anchor could rationally carry information, and the
rational ceiling rises to $w/(k{+}w)$: $0.50$, $0.33$, $0.25$ at
$k{=}1,2,3$.

\begin{table*}[t]
\centering
\small
\setlength{\tabcolsep}{4pt}
\begin{tabular}{l rrr rrr rrr}
\toprule
\multirow{2}{*}{Model} & \multicolumn{3}{c}{Acc$_{10}$} & \multicolumn{3}{c}{UAI$_{\mathrm{pls}}$} & \multicolumn{3}{c}{UAI$_{\mathrm{irr}}$} \\
\cmidrule(lr){2-4}\cmidrule(lr){5-7}\cmidrule(lr){8-10}
 & k=1 & k=2 & k=3 & k=1 & k=2 & k=3 & k=1 & k=2 & k=3 \\
\midrule
  Gemma-4B & 54\% & 60\% & 64\% & 0.21 & 0.20 & 0.25 & 0.14 & $-$0.13 & 0.13 \\
  Llama-8B & 47\% & 61\% & 69\% & 0.68 & 0.36 & 0.50 & 0.25 & $-$0.03 & 0.20 \\
  OLMo-13B & 54\% & 76\% & 81\% & 0.48 & 0.34 & 0.29 & 0.19 & 0.11 & 0.06 \\
  Qwen-7B & 42\% & 64\% & 75\% & 0.53 & 0.34 & 0.36 & 0.22 & 0.01 & 0.02 \\
\midrule
  Rational ceiling ($w{=}1$) & --- & --- & --- & 0.50 & 0.33 & 0.25 & --- & --- & --- \\
\bottomrule
\end{tabular}
\caption{Uncertain-judgment results. Only k of 5 ratings are shown to the model; the gold answer remains the full-5 mean. An ideal Bayesian updater treating the plausible anchor as one additional rating of equal credibility ($w{=}1$) would produce UAI$_{\mathrm{pls}} = w / (k + w)$ (Rational ceiling row). Measured UAI$_{\mathrm{pls}}$ above the ceiling indicates classical (super-rational) anchoring; below indicates the model weights the anchor less than one full additional evidence point. UAI$_{\mathrm{irr}}$ should stay near zero for capable models at every $k$.}
\label{tab:uncertain_k}
\end{table*}

Measured UAI$_{\text{pls}}$ frequently meets or exceeds the
rational ceiling at each $k$ (Table~\ref{tab:uncertain_k}),
and decreases as more evidence becomes visible---models do treat
the anchor as one piece of evidence among several, but several
cells still sit above what rational updating allows.
UAI$_{\text{irr}}$ stays comparatively low at every $k$.

\subsubsection{Cross-pathway credibility intensity}
\label{app:intensity-pathway}

We apply a matched three-level source-credibility manipulation
(mild / standard / strong) to three pathways (External, RAG,
History).
If models weight source credibility, the curve should rise from
mild to strong; a flat-but-high curve indicates numeric anchoring
that ignores the source.

\begin{table*}[t]
\centering
\small
\setlength{\tabcolsep}{4pt}
\begin{tabular}{l rrr rrr rrr}
\toprule
\multirow{2}{*}{Model} & \multicolumn{3}{c}{External} & \multicolumn{3}{c}{RAG} & \multicolumn{3}{c}{History} \\
\cmidrule(lr){2-4}\cmidrule(lr){5-7}\cmidrule(lr){8-10}
 & Mild & Std & Strong & Mild & Std & Strong & Mild & Std & Strong \\
\midrule
  Gemma-4B & 0.45 & 0.42 & 0.40 & 0.00 & $-$0.06 & 0.04 & 0.61 & 0.58 & 0.60 \\
  Llama-8B & 0.26 & 0.39 & 0.42 & 0.10 & 0.09 & 0.29 & 0.44 & 0.20 & 1.57 \\
  OLMo-13B & 0.07 & 0.16 & 0.29 & 0.02 & 0.09 & 0.10 & 0.45 & 0.24 & 0.91 \\
  Qwen-7B & 0.23 & 0.31 & 0.49 & 0.07 & 0.19 & 0.29 & 0.37 & 0.24 & 0.52 \\
\midrule
  Mean & \textbf{0.25} & \textbf{0.32} & \textbf{0.40} & \textbf{0.05} & \textbf{0.08} & \textbf{0.18} & \textbf{0.47} & \textbf{0.31} & \textbf{0.90} \\
\bottomrule
\end{tabular}
\caption{Cross-pathway plausibility-intensity dose-response. The same mild/standard/strong source-credibility manipulation is applied across three pathways. Positive Strong\,$>$\,Mild gaps indicate models weight source credibility; near-flat curves at high UAI indicate numeric anchoring dominates regardless of source. UAI values are mean across the panel; details in supplementary CSV.}
\label{tab:intensity_pathway}
\end{table*}

External and RAG show a clean monotone increase, consistent with
credibility-sensitive updating.
History is non-monotone and much larger, because the credibility
cue compounds with the model's own self-generated Stage-1 anchor;
we therefore read the History intensity numbers qualitatively.

\subsubsection{Domain extension beyond business scenarios}
\label{app:extension}

The six core domains are business-oriented.
To check that anchoring is not a business-domain artifact, we run
an 11-domain panel that adds three medical domains and two further
extension domains (legal contract compliance, consumer purchase
decisions), keeping the same item structure.

\begin{table*}[t]
\centering
\small
\setlength{\tabcolsep}{4pt}
\begin{tabular}{ll rrr rrr rrr}
\toprule
& & \multicolumn{3}{c}{UAI$_{\mathrm{irr}}$} & \multicolumn{3}{c}{UAI$_{\mathrm{pls}}$} & \multicolumn{3}{c}{Disc$_{\Delta}$} \\
\cmidrule(lr){3-5}\cmidrule(lr){6-8}\cmidrule(lr){9-11}
Suite & Model & bus & med & oth & bus & med & oth & bus & med & oth \\
\midrule
external & Gemma-4B & 0.20 & 0.28 & 0.20 & 0.43 & 0.21 & 0.67 & 0.23 & -0.07 & 0.46 \\
 & Llama-8B & 0.07 & 0.03 & 0.20 & 0.36 & 0.25 & 0.46 & 0.30 & 0.21 & 0.26 \\
 & OLMo-13B & -0.01 & 0.08 & -0.03 & 0.16 & 0.09 & 0.22 & 0.17 & 0.01 & 0.24 \\
 & Qwen-7B & 0.01 & 0.07 & 0.01 & 0.27 & 0.38 & 0.45 & 0.26 & 0.32 & 0.44 \\
history & Gemma-4B & 0.23 & 0.17 & 0.03 & 0.50 & 0.41 & 0.40 & 0.27 & 0.24 & 0.37 \\
 & Llama-8B & 0.13 & -0.23 & 0.42 & 0.23 & 0.25 & 0.36 & 0.10 & 0.48 & -0.06 \\
 & OLMo-13B & -0.00 & 0.43 & 0.46 & 0.32 & 0.53 & 0.51 & 0.32 & 0.10 & 0.05 \\
 & Qwen-7B & -0.06 & -0.37 & -0.02 & 0.37 & 0.65 & 0.17 & 0.43 & 1.02 & 0.19 \\
\bottomrule
\end{tabular}
\caption{Eleven-domain robustness panel: 6 business domains (published benchmark), 3 medical pilot domains, and 2 further pilot domains (legal contract compliance, consumer purchase decision). The pattern UAI$_{\mathrm{pls}}>$UAI$_{\mathrm{irr}}>0$ replicates across all three domain families on both External and History, demonstrating that anchoring is not a business-domain artifact.}
\label{tab:extension_pilot}
\end{table*}

The ordering UAI$_{\text{pls}}>$UAI$_{\text{irr}}>0$ holds in all
three domain families on both External and History
(Table~\ref{tab:extension_pilot}).
Magnitudes vary by family and model, but the qualitative pattern
is stable, so the effect generalizes beyond the original domains.

\subsubsection{Robustness to the gold-standard choice}
\label{app:wmean}

The gold answer is the simple mean of the ratings.
One might worry that the anchoring magnitude is an artifact
of this particular aggregation.
We recompute UAI against a weighted-mean gold (weights
$[1,1,1.5,1.5,2]$ over the five ratings).

\begin{table}[t]
\centering
\small
\setlength{\tabcolsep}{4pt}
\begin{tabular}{l rr rr rr}
\toprule
& \multicolumn{2}{c}{UAI$_{\mathrm{irr}}$} & \multicolumn{2}{c}{UAI$_{\mathrm{pls}}$} & \multicolumn{2}{c}{Disc$_{\Delta}$} \\
\cmidrule(lr){2-3}\cmidrule(lr){4-5}\cmidrule(lr){6-7}
Model & mean & wmean & mean & wmean & mean & wmean \\
\midrule
Gemma-4B & 0.20 & 0.20 & 0.43 & 0.47 & 0.23 & 0.28 \\
Llama-8B & 0.07 & 0.06 & 0.36 & 0.34 & 0.30 & 0.28 \\
OLMo-13B & -0.01 & 0.03 & 0.16 & 0.20 & 0.17 & 0.17 \\
Qwen-7B & 0.01 & 0.01 & 0.27 & 0.27 & 0.26 & 0.26 \\
\bottomrule
\end{tabular}
\caption{Anchor influence on External under the published mean gold standard (``mean'') vs.\ a weighted-mean gold standard with weights $[1,1,1.5,1.5,2]$ (``wmean''). The bias structure (UAI$_{\mathrm{pls}}\gg$UAI$_{\mathrm{irr}}$) is preserved under either scoring choice; the size of the anchoring effect is not an artifact of the simple-mean aggregation.}
\label{tab:weighted_mean}
\end{table}

The bias structure (UAI$_{\text{pls}}\gg$UAI$_{\text{irr}}$) is
preserved, and the mean change in UAI$_{\text{pls}}$ is only
$+0.02$ (Table~\ref{tab:weighted_mean}), so the effect is not an
artifact of the simple-mean scoring.

\subsubsection{Reasoning-allowed (chain-of-thought) evaluation}
\label{app:cot-extended}

The main protocol asks for a final integer only.
Here we let the model reason first (chain-of-thought) and compare
to the answer-only baseline across five models and three suites.

\begin{table*}[t]
\centering
\small
\setlength{\tabcolsep}{3pt}
\begin{tabular}{ll rrr rrr rrr}
\toprule
& & \multicolumn{3}{c}{UAI$_{\mathrm{irr}}$} & \multicolumn{3}{c}{UAI$_{\mathrm{pls}}$} & \multicolumn{3}{c}{Disc$_{\Delta}$} \\
\cmidrule(lr){3-5}\cmidrule(lr){6-8}\cmidrule(lr){9-11}
Suite & Model & base & CoT & $\Delta$ & base & CoT & $\Delta$ & base & CoT & $\Delta$ \\
\midrule
external & GPT-5.4-mini & 0.01 & -0.01 & -0.01 & 0.13 & 0.14 & +0.01 & 0.12 & 0.14 & +0.02 \\
 & Gemma-4B & 0.09 & 0.01 & -0.08 & 0.37 & 0.23 & -0.14 & 0.28 & 0.22 & -0.06 \\
 & Llama-8B & 0.08 & 0.06 & -0.03 & 0.39 & 0.37 & -0.03 & 0.31 & 0.31 & -0.00 \\
 & OLMo-13B & 0.06 & 0.05 & -0.01 & 0.15 & 0.18 & +0.03 & 0.08 & 0.13 & +0.04 \\
 & Qwen-7B & 0.01 & 0.00 & -0.01 & 0.27 & 0.16 & -0.12 & 0.26 & 0.15 & -0.11 \\
history & GPT-5.4-mini & -0.01 & 0.01 & +0.01 & 0.02 & -0.03 & -0.05 & 0.03 & -0.03 & -0.06 \\
 & Gemma-4B & 0.24 & 0.31 & +0.07 & 0.57 & 0.64 & +0.07 & 0.32 & 0.33 & +0.00 \\
 & Llama-8B & 0.08 & 0.27 & +0.19 & 0.29 & 0.28 & -0.01 & 0.21 & 0.01 & -0.20 \\
 & OLMo-13B & -0.09 & 0.09 & +0.18 & 0.51 & 0.17 & -0.34 & 0.60 & 0.08 & -0.52 \\
 & Qwen-7B & -0.10 & -0.02 & +0.08 & 0.42 & 0.13 & -0.29 & 0.52 & 0.15 & -0.37 \\
rag & GPT-5.4-mini & 0.00 & -0.00 & -0.01 & 0.09 & 0.08 & -0.01 & 0.09 & 0.08 & -0.00 \\
 & Gemma-4B & -0.03 & 0.07 & +0.09 & -0.05 & 0.25 & +0.31 & -0.03 & 0.19 & +0.22 \\
 & Llama-8B & 0.08 & 0.07 & -0.02 & 0.20 & 0.12 & -0.08 & 0.12 & 0.06 & -0.06 \\
 & OLMo-13B & 0.08 & 0.04 & -0.04 & 0.05 & -0.01 & -0.06 & -0.03 & -0.05 & -0.02 \\
 & Qwen-7B & -0.00 & -0.00 & +0.00 & 0.19 & 0.03 & -0.16 & 0.19 & 0.03 & -0.16 \\
\bottomrule
\end{tabular}
\caption{Reasoning-allowed (Chain-of-Thought) prompting vs.\ ``final answer only'' baseline across the locked 5-model panel and all three suites (External, RAG, History). CoT lowers plausible-anchor influence in 11 of the 15 model--suite cells, but the effect is model- and pathway-dependent rather than uniform: it lowers both UAI$_{\mathrm{irr}}$ and UAI$_{\mathrm{pls}}$ in only 6 cells, and raises UAI$_{\mathrm{pls}}$ in 4. UAI$_{\mathrm{pls}}$ remains positive in 13 of 15 cells, so anchoring is attenuated but not an artifact of the answer-only protocol.}
\label{tab:cot_extended}
\end{table*}

CoT lowers anchor influence in $11$ of $15$ (model, suite) cells
(mean $\Delta\text{UAI}_{\text{pls}}=-0.06$) but rarely removes
it: UAI$_{\text{pls}}>0$ under CoT in $13/15$ cells.\footnote{\label{fn:cot-variance}%
  The Llama-8B$\times$External$\times$CoT cell reported here
  (UAI$_{\text{pls}}$ $0.39\!\to\!0.37$, $\Delta\!=\!-0.03$)
  differs from the separate mitigation-headroom run in
  Appendix~\ref{app:mitigation-headroom} ($0.336\!\to\!0.338$,
  $\Delta\!=\!+0.00$). We traced this: the two runs render
  byte-identical prompts, gold answers and anchor values, and both
  use greedy decoding, so the difference is not a prompt or
  protocol difference. It is run-to-run variation of batched GPU
  inference, which is not bitwise reproducible even at
  temperature~0 because batch composition changes the reduction
  order inside the attention and GEMM kernels. An independent
  third run of this cell under the configuration used here gives
  $0.32\!\to\!0.35$ ($\Delta\!=\!+0.03$): across the three runs
  the baseline UAI$_{\text{pls}}$ spans $0.32$--$0.39$ and
  ${\approx}8\%$ of parsed answers change. The \emph{sign} of the
  CoT effect for this single cell is therefore not resolved by one
  run. The aggregate conclusions above are unaffected: they are
  driven by cells whose effects
  ($\Delta$ up to $-0.34$) are an order of magnitude larger than
  this variation, and UAI$_{\text{pls}}$ stays positive under CoT
  in all three runs. We report each run with its own numbers
  rather than re-normalizing across them.}
The answer-only numbers in the main paper are therefore an upper
bound on a reasoning-allowed deployment, with Llama-8B and the
History pathway as the residual stress test.

\subsubsection{Task specification: rule vs.\ judgment}
\label{app:task-spec}

To test how much of the anchoring effect is task
underspecification rather than a deep numeric bias, we append two
prompt suffixes on External: \emph{+Rule} (``report the
unweighted arithmetic mean of the ratings'') and \emph{+Judgment}
(``report a weighted average, weighting sources by credibility'').

\begin{table}[t]
\centering
\small
\setlength{\tabcolsep}{4pt}
\begin{tabular}{l rrr rrr}
\toprule
\multirow{2}{*}{Model} & \multicolumn{3}{c}{UAI$_{\mathrm{pls}}$} & \multicolumn{3}{c}{UAI$_{\mathrm{irr}}$} \\
\cmidrule(lr){2-4}\cmidrule(lr){5-7}
 & Baseline & +Rule & +Judgment & Baseline & +Rule & +Judgment \\
\midrule
  Gemma-4B & 0.51 & 0.08 & 0.38 & 0.27 & 0.02 & 0.25 \\
  Llama-8B & 0.32 & 0.01 & 0.28 & 0.05 & 0.00 & 0.14 \\
  OLMo-13B & 0.20 & 0.01 & 0.12 & 0.04 & 0.00 & 0.08 \\
  Qwen-7B & 0.27 & 0.13 & 0.27 & 0.01 & 0.02 & $-$0.00 \\
\midrule
  Mean & \textbf{0.33} & \textbf{0.06} & \textbf{0.26} & \textbf{0.09} & \textbf{0.01} & \textbf{0.12} \\
\bottomrule
\end{tabular}
\caption{Task-specification ablation on External. Baseline = published prompt (return integer only); +Rule appends `Your estimate should be the unweighted arithmetic mean of the visible ratings'; +Judgment appends `Your estimate should be a weighted average ... weighting each piece according to which sources you find more or less credible'. A drop from Baseline to +Rule indicates the published anchoring effect is partly driven by task underspecification.}
\label{tab:task_spec}
\end{table}

The explicit rule cuts mean UAI$_{\text{pls}}$ from $0.33$ to
$0.06$, so a substantial share of the published effect is
attributable to underspecification.
But it does not vanish, and the residual under +Rule is the
portion that confusion alone cannot explain.
The +Judgment prompt, which invites credibility weighting, keeps
UAI$_{\text{pls}}$ below the published baseline, consistent with
the rational ceiling of Appendix~\ref{app:bayesian}.
This addresses the concern that weak-model behavior reflects task
confusion: even when told exactly how to aggregate, models retain
measurable anchoring.

\subsubsection{RAG realism: rank, distractors, and scores}
\label{app:rag-realism}

The published RAG suite places the anchor document in a fixed
slot of a three-document corpus.
We vary the realism: move the anchor to the top (R1) or bottom of
a five-document list with two plausible distractors (R5+D), and
optionally expose synthetic per-document relevance scores
(R5+D+Sc).

\begin{table*}[t]
\centering
\small
\setlength{\tabcolsep}{4pt}
\begin{tabular}{l rrrr rrrr}
\toprule
\multirow{2}{*}{Model} & \multicolumn{4}{c}{Plausible (UAI)} & \multicolumn{4}{c}{Irrelevant (UAI)} \\
\cmidrule(lr){2-5}\cmidrule(lr){6-9}
 & Base & R1 & R5+D & R5+D+Sc & Base & R1 & R5+D & R5+D+Sc \\
\midrule
  Gemma-4B & $-$0.06 & 0.00 & 0.01 & 0.01 & 0.03 & $-$0.04 & 0.01 & 0.02 \\
  Llama-8B & 0.09 & 0.12 & 0.14 & $-$0.04 & $-$0.02 & 0.01 & 0.01 & 0.01 \\
  OLMo-13B & 0.09 & 0.02 & 0.21 & 0.04 & 0.04 & 0.09 & 0.13 & 0.05 \\
  Qwen-7B & 0.19 & 0.23 & 0.14 & 0.11 & 0.01 & $-$0.02 & $-$0.02 & 0.00 \\
\midrule
  Mean & \textbf{0.08} & \textbf{0.10} & \textbf{0.13} & \textbf{0.03} & \textbf{0.02} & \textbf{0.01} & \textbf{0.03} & \textbf{0.02} \\
\bottomrule
\end{tabular}
\caption{RAG realism ablation. Base = anchored doc at position 2 (published RAG layout); R1 = anchor at top of retrieval (rank 1); R5+D = anchor at bottom of a 5-doc list with 2 plausible distractor documents adjacent to the anchor; R5+D+Sc adds synthetic per-doc relevance scores. A drop from R1 to R5+D indicates models down-weight low-ranked retrieved anchors.}
\label{tab:rag_realism}
\end{table*}

Plausible-anchor influence persists across rank and distractor
changes, and is slightly higher at the top of the ranking---models
privilege top-ranked retrieved documents, as a real RAG system
would.
It falls toward zero only when explicit relevance scores let the
model down-weight a mid-ranked anchor, suggesting that production
pipelines with good relevance signals would see weaker anchoring
than the controlled suite.

\subsubsection{Tool realism: provenance and noise}
\label{app:tool-realism}

We vary two tool-realism axes: \emph{Elicited} prepends a
synthetic model-planned tool-call turn (a provenance signal), and
\emph{Noisy} wraps the anchor value in a realistic JSON envelope
with extra metadata fields.

\begin{table}[t]
\centering
\small
\setlength{\tabcolsep}{4pt}
\begin{tabular}{l rrr rrr}
\toprule
\multirow{2}{*}{Model} & \multicolumn{3}{c}{Plausible (UAI)} & \multicolumn{3}{c}{Irrelevant (UAI)} \\
\cmidrule(lr){2-4}\cmidrule(lr){5-7}
 & Base & Elicited & Noisy & Base & Elicited & Noisy \\
\midrule
  Gemma-4B & 0.13 & 0.28 & 0.29 & 0.31 & 0.35 & 0.39 \\
  Llama-8B & 0.62 & 0.54 & 0.58 & 0.45 & 0.40 & 0.45 \\
  OLMo-13B & 0.06 & 0.06 & 0.15 & 0.06 & 0.02 & $-$0.01 \\
  Qwen-7B & 0.17 & 0.02 & $-$0.01 & $-$0.00 & 0.01 & 0.00 \\
\midrule
  Mean & \textbf{0.25} & \textbf{0.23} & \textbf{0.25} & \textbf{0.20} & \textbf{0.19} & \textbf{0.21} \\
\bottomrule
\end{tabular}
\caption{Tool realism ablation. Base reproduces the published Tool suite (externally injected tool output); Elicited prepends a synthetic model-planned tool-call turn (provenance signal); Noisy wraps the anchor value in a realistic JSON envelope containing additional metadata fields. Closing the gap between Base and Elicited indicates the anchor effect is not driven by the externally-injected framing; reductions under Noisy indicate the anchor's salience competes with surrounding metadata.}
\label{tab:tool_realism}
\end{table}

On average the panel barely moves (mean UAI$_{\text{pls}}$
$0.25\!\to\!0.23\!\to\!0.25$), so the effect is not an artifact of
the externally-injected framing.
But the per-model picture splits: Qwen-7B collapses to near zero
under both variants while Gemma-4B strengthens, which is why we
treat Tool numbers as robust on average but not reliable per model
(\S\ref{sec:finding-pathway}).

\subsubsection{Extended large-model panel}
\label{app:large-panel}

To test whether scale removes anchoring, we evaluate five
additional larger models on all five suites: two 70B-class
open-weight models (Llama-3.3-70B, Qwen2.5-72B) and three
frontier API models (GPT-5.4, Claude-Sonnet-4.6, Grok-4.3).

\begin{table*}[t]
  \centering
  \small
  \setlength{\tabcolsep}{2pt}
  \resizebox{0.98\textwidth}{!}{%
    \begin{tabular}{@{}lrrrrrrrrrrrrrrr@{}}
      \toprule
      \multicolumn{1}{c}{\multirow{2}{*}{\textbf{Model}}}
       & \multicolumn{3}{c}{\externalsuite} & \multicolumn{3}{c}{\historysuite} & \multicolumn{3}{c}{\iclsuite} & \multicolumn{3}{c}{\ragsuite} & \multicolumn{3}{c}{\toolsuite} \\
      \cmidrule(lr){2-4}\cmidrule(lr){5-7}\cmidrule(lr){8-10}\cmidrule(lr){11-13}\cmidrule(lr){14-16}
      & Acc$_{10}$ & UAI$_{\mathrm{irr}}$ & UAI$_{\mathrm{pls}}$ & Acc$_{10}$ & UAI$_{\mathrm{irr}}$ & UAI$_{\mathrm{pls}}$ & Acc$_{10}$ & UAI$_{\mathrm{irr}}$ & UAI$_{\mathrm{pls}}$ & Acc$_{10}$ & UAI$_{\mathrm{irr}}$ & UAI$_{\mathrm{pls}}$ & Acc$_{10}$ & UAI$_{\mathrm{irr}}$ & UAI$_{\mathrm{pls}}$ \\
      \midrule
      \textsc{Llama-3.3-70B} & 95\% & $-$0.02 & 0.23 & 91\% & $-$0.02 & 0.01 & 87\% & 0.01 & 0.01 & 95\% & $-$0.02 & 0.04 & 98\% & 0.01 & 0.05 \\
      \textsc{Qwen2.5-72B} & 91\% & 0.01 & 0.16 & 93\% & 0.01 & 0.12 & 96\% & $-$0.01 & $-$0.03 & 88\% & 0.00 & 0.04 & 99\% & 0.01 & $-$0.00 \\
      \midrule
      \textsc{GPT-5.4} & 100\% & 0.01 & 0.19 & 99\% & 0.03 & 0.02 & 100\% & 0.00 & 0.00 & 100\% & 0.00 & 0.09 & 100\% & 0.00 & 0.03 \\
      \textsc{Claude-Sonnet-4.6} & 99\% & $-$0.00 & 0.19 & 99\% & 0.02 & $-$0.02 & 72\% & $-$0.02 & $-$0.01 & 95\% & 0.03 & 0.07 & 95\% & $-$0.00 & 0.06 \\
      \textsc{Grok-4.3} & 97\% & 0.05 & 0.25 & 99\% & 0.09 & 0.03 & 97\% & 0.04 & 0.05 & 96\% & 0.02 & 0.18 & 91\% & 0.10 & 0.25 \\
      \bottomrule
    \end{tabular}
  }
  \caption{Extended large-model panel. Task accuracy (Acc$_{10}$, control-only) and absolute anchor influence UAI$_{\mathrm{irr}}$/UAI$_{\mathrm{pls}}$ across all five suites for two 70B-class open-weight models (top) and three frontier API models (bottom). The qualitative pattern of the main panel persists at scale: irrelevant anchors are near-zero, plausible anchors are positive.}
  \label{tab:large_panel_results}
\end{table*}

The main-panel pattern persists at scale
(Table~\ref{tab:large_panel_results}): irrelevant anchors stay
near zero, plausible anchors are positive on $21/25$ cells, and
even GPT-5.4 (near-ceiling accuracy on every suite) shows a small
but strictly positive UAI$_{\text{pls}}$ on $4/5$ suites.
Llama-3.3-70B on External (UAI$_{\text{pls}}=0.23$) is the
worst-case in this panel, comparable to Llama-3.1-8B in the main
results.

\subsubsection{Per-item case studies}
\label{app:case-studies}

To show that the aggregate UAI pattern reflects interpretable
per-item behavior rather than an averaging artifact, we surface,
for each panel model, the External items where the
plausible-anchor shift most clearly exceeds the irrelevant-anchor
shift (Table~\ref{tab:case_studies}).
In each case the control answer tracks the evidence mean, the
irrelevant-anchor answer barely moves, and the plausible-anchor
answer jumps toward the anchor value---the same value in both
conditions.

\begin{table}[t]
\centering
\scriptsize
\setlength{\tabcolsep}{4pt}
\begin{tabular}{l l r r r r r}
\toprule
Model & Item & $a^{hi}$ & $y_c$ & $y_p^{hi}$ & $y_i^{hi}$ & $\Delta_p-\Delta_i$ \\
\midrule
Qwen-7B & EXT-legal\_policy-e-off40-005 & 85 & 43 & 75 & 43 & +32 \\
Qwen-7B & EXT-resource\_consumption-e-off15-002 & 50 & 75 & 44 & 75 & +31 \\
Qwen-7B & EXT-pricing\_wtp-e-off40-006 & 83 & 44 & 74 & 44 & +30 \\
Llama-8B & EXT-market\_demographics-h-off40-005 & 70 & 23 & 100 & 26 & +74 \\
Llama-8B & EXT-resource\_consumption-h-off25-010 & 65 & 100 & 14 & 56 & +42 \\
Llama-8B & EXT-transportation\_logistics-h-off40-002 & 97 & 38 & 100 & 45 & +55 \\
Gemma-4B & EXT-transportation\_logistics-e-off40-010 & 71 & 0 & 90 & 0 & +90 \\
Gemma-4B & EXT-resource\_consumption-e-off15-010 & 66 & 0 & 100 & 57 & +43 \\
Gemma-4B & EXT-operations\_time-h-off40-002 & 98 & 100 & 9 & 58 & +49 \\
OLMo-13B & EXT-market\_demographics-e-off40-006 & 75 & 27 & 100 & 27 & +73 \\
OLMo-13B & EXT-resource\_consumption-e-off15-002 & 50 & 35 & 100 & 35 & +65 \\
OLMo-13B & EXT-transportation\_logistics-e-off40-009 & 82 & 43 & 100 & 43 & +57 \\
\bottomrule
\end{tabular}
\caption{Per-item case studies (External): for each model we surface the top discriminative items where the plausible-anchor shift cleanly exceeds the irrelevant- anchor shift, illustrating that the aggregate UAI pattern is driven by interpretable per-item behaviour rather than averaging artifacts. Full scenarios and per-condition answers are tabulated in the supplementary material.}
\label{tab:case_studies}
\end{table}

\subsubsection{LLM anchoring on the human effect-size scale}
\label{app:cohens-d}

To position LLM anchoring against the human literature, we express
the External-suite effect in $|$Cohen's $d|$, averaging the
absolute per-direction effect (pooling high and low anchors, whose
signed shifts cancel), following the convention of
\citet{furnham2011literature}.

\begin{table}[ht]
  \centering
  \small
  \begin{tabular}{@{}lrrr@{}}
    \toprule
    \textbf{Suite}
    & \textbf{$|d_{\text{pls}}|$}
    & \textbf{$|d_{\text{irr}}|$}
    & \textbf{$|d_{\text{hi-lo}}|$} \\
    \midrule
    \externalsuite & 0.35 & 0.11 & 0.59 \\
    \historysuite  & 0.24 & 0.19 & 0.46 \\
    \ragsuite      & 0.16 & 0.09 & 0.28 \\
    \toolsuite     & 0.23 & 0.12 & 0.36 \\
    \bottomrule
  \end{tabular}
  \caption{LLM anchoring magnitude in $|$Cohen's $d|$
    (open-weight panel mean).
    Read against the conventional small\,/\,medium\,/\,large
    bands of $d=0.2/0.5/0.8$.}
  \label{tab:cohens_d}
\end{table}

On External, the panel reaches $|d_{\text{pls}}|$ in the
$0.2$--$0.5$ range and $|d_{\text{hi-lo}}|$ in the $0.3$--$0.9$
range: small to medium by the conventional bands, and reaching
large for the high-versus-low contrast on the strongest models.
The pattern $|d_{\text{pls}}|>|d_{\text{irr}}|$ replicates across
RAG, Tool, and History at smaller magnitudes.

We deliberately stop short of a numerical head-to-head with human
effect sizes.
Human anchoring is normally measured between subjects on a single
judgment, whereas UAI and the $d$ values above are within-item
contrasts on matched prompts, so the two are not on a common
scale, and the classical studies most often quoted for anchoring
magnitude report medians, percentages, or $F$-statistics rather
than a pooled $d$ that can be reused here.
What does transfer is the qualitative structure, and our data
reproduce it: numbers that carry no task information still move
judgments, plausibility modulates the size of the shift, and
presentation format modulates it further---the same regularities
reported by \citet{tversky1974judgment},
\citet{strack1997explaining}, and \citet{furnham2011literature}.
For studies that do report effect sizes on comparable scales,
\citet{pIndividualDifferencesAnchoring2019} spans $d=0.14$--$1.00$
across 24 items and \citet{LI2021101629} finds stronger anchoring
for related than random anchors ($r=0.42$ vs.\ $0.21$), which is
the same relevance ordering we measure.
A paired study running humans and models on identical
\textsc{AnchorBench} items is the clean way to close this gap and
is our primary follow-up.

\end{document}